\documentclass[11pt,onecolumn]{style}
\usepackage{placeins}

\title{MM-ABC: Towards Generalist Mobile Manipulation \\ via Seeing, Coordinating and Imagining}

\author{%
  \textbf{Qiwei Liang}\authmark{1,2,*,\textdaggerdbl}\quad
  \textbf{Guangyu Chen}\authmark{1,3,*}\quad
  \textbf{Shaolong Zhu}\authmark{1,3,*}\quad
  \textbf{Zikuan Xiao}\authmark{1}\quad
  \textbf{Jinxuan Lu}\authmark{1,2}\\
  \textbf{Yifan Xie}\authmark{3}\quad
  \textbf{Renjing Xu}\authmark{2,\textdagger}\quad
  \textbf{Wenbo Ding}\authmark{1,3,\textdagger}\quad
  \textbf{Tianxing Chen}\authmark{1,4,\textdagger}
}

\affiliation{%
  \authmark{1} Xspark AI \quad
  \authmark{2} The Hong Kong University of Science and Technology (Guangzhou) \\
  \authmark{3} Tsinghua University \quad
  \authmark{4} The University of Hong Kong
}

\correspondence{%
  \authmark{*} Equal Contribution \quad
  \authmark{\textdaggerdbl} Project Lead \quad
  \authmark{\textdagger} Corresponding Authors
}

\homepage{%
  \href{https://mm-abc.github.io/}{\textbf{Project Page}} \quad \textcolor{mmabcLightMuted}{\textbullet} \quad
  \href{https://github.com/MM-ABC/MM-ABC}{\textbf{Codebase \& Model Zoo}}
}

\date{September 2026}

\hypersetup{
  pdftitle={MM-ABC: Towards Generalist Mobile Manipulation via Seeing, Coordinating and Imagining},
  pdfsubject={Mobile Manipulation via Arm-Base Collaboration},
  pdfkeywords={MM-ABC, Mobile Manipulation, Arm-Base Collaboration, Embodied AI}
}

\begin{document}
\maketitle

% ==============================================================================
% Abstract
% ==============================================================================
\begin{abstract}

  Mobile manipulation extends robot interaction beyond a fixed kinematic workspace by making the reachable region itself controllable.
  This flexibility introduces two central challenges: spatially grounded perception under continuous ego-motion and coordinated control of heterogeneous arm and base actions.
  Existing approaches strengthen geometry through explicit 3D representations or predictive world models, and often decouple mobility and manipulation into separate action streams.
  We argue that effective mobile manipulation requires not only decoupling, but also representations that support efficient cross-stream collaboration.
  We present \textbf{MM-ABC}, a foundation model built around \textbf{Seeing, Coordinating, and Imagining Arm--Base Collaboration}.
  MM-ABC combines sparse multi-level VLM features for spatial perception; a training-only future branch that uses world imagination and geometric intent as extra supervision, strengthening perception and manipulation-intent prediction and improving the overall learning signal; and MM-APT, which coordinates separate manipulation and mobility streams through masked joint attention and clean-action $x$-prediction.
  In controlled ablations, replacing clean-action prediction with velocity prediction lowers success on RoboCasa365 composite-seen tasks from 32.8\% to 29.2\%, and removing future supervision or multilevel conditioning causes larger drops.
  We pretrain MM-ABC on \textbf{5,000+ hours} of heterogeneous robot data spanning \textbf{400K+ episodes, 12 datasets, and 17 embodiments}.
  Experiments cover EBench, RoboCasa365, ManiSkill-HAB, LIBERO, LIBERO-Plus, and real-world mobile manipulation.
  MM-ABC achieves 44.71\% success on EBench, 61.2\% on RoboCasa365, 99.1\% on LIBERO, 82.8\% on LIBERO-Plus without perturbation training, and 83\% mean success on five real-world tasks.

  
  \end{abstract}

% ==============================================================================
% Modular Sections
% ==============================================================================
\section{Introduction}

Robot manipulation ultimately requires bringing the robot's joints and end effectors into suitable configurations to interact with objects in the physical world.
For a fixed-base manipulator, however, this interaction is fundamentally bounded by its kinematic workspace: once a target lies outside the reachable region, even a capable policy cannot interact with it.
A mobile base changes this constraint.
By actively repositioning the robot, it turns the reachable workspace from a fixed property into a controllable variable, extending manipulation from local tabletops to rooms and large-scale environments.
From this perspective, \emph{mobile manipulation is not merely navigation followed by manipulation; it is manipulation over a dynamically reconfigurable workspace whose extent the policy itself controls}.

This expanded workspace introduces two central challenges.
First, \textbf{spatial perception becomes dynamic}: ego-motion continuously changes viewpoints and reference frames, while interaction still requires precise relative geometry among the robot, target, and scene.
Second, \textbf{arm--base actions must be coordinated}.
The base performs large-scale repositioning, whereas the manipulator executes fine-grained interaction.
Their dynamics and temporal roles differ substantially, yet they are inherently coupled: base motion determines what the arm can reach, while the intended manipulation determines where the base should move.

Existing work addresses the spatial challenge through point clouds, depth, geometric tokens, dynamic-aware 3D representations, spatial memories, or auxiliary geometric objectives~\citep{liu2026incom,zhu2026geohat,liang2026afro,tu2026sgvla}.
Predictive models further learn future visual or geometric representations to improve action learning~\citep{zheng2025flare,chen2026abotm05,li2026wam4d}.
While effective, explicit reconstruction or dense future generation can introduce substantial modeling and inference cost.
We instead ask whether \emph{world imagination} and \emph{geometric intent} can serve purely as extra training supervision: denser geometric targets that improve perception and manipulation-intent prediction, and thereby the overall learning signal, without requiring a world model at deployment.

The coordination problem raises a complementary question.
Recent methods increasingly decouple mobility and manipulation through separate action branches, subsystem-specific perception, or structured action spaces~\citep{chen2025acdit,liu2026incom,zhu2026geohat,chen2026mopa,chen2026abotm05}.
However, \emph{decoupling alone does not guarantee collaboration}.
Once arm and base are represented by different streams, the key design question becomes how these streams should exchange information while preserving their own physical structure and distinct control semantics.

This motivates us to revisit the output parameterization of flow-based action generation.
Under rectified flow, clean-action $x$-prediction and velocity $v$-prediction are algebraically equivalent, but they need not impose the same representational burden on a finite-width multi-stream expert.
A velocity head must retain and cancel high-dimensional flow noise to recover the endpoint, whereas a clean-action head can map directly toward the action manifold.
In a controlled synthetic study with a known Bayes-optimal denoiser, $x$-prediction retains far less injected noise in its endpoint estimates and denoises more accurately at high noise.
Under few-step sampling, it also realizes the prescribed arm--body task allocation more accurately.
These observations motivate clean-action prediction in our two-stream action expert, whose streams share a finite width.

Building on these observations, we introduce \textbf{MM-ABC}, organized around three principles: \textbf{Seeing, Coordinating, and Imagining}.
For \textbf{seeing}, MM-ABC uses a sparse \emph{DeepStack} interface that injects multi-level VLM features into the action expert, preserving complementary fine-grained and semantic information.
For \textbf{imagining}, learnable future queries predict geometry-rich representations at sparse horizons under a frozen geometric teacher, capturing workspace evolution and the geometric intent the policy should realize.
This supervision densifies training signals for current-scene perception and manipulation-intent prediction, while future representations never access action tokens and the branch is removed at deployment.
For \textbf{coordinating}, we introduce the \textbf{Mobile Manipulation Action Prediction Transformer (MM-APT)}, which maintains separate manipulation and mobile/body streams while enabling cross-stream interaction through masked joint attention.
MM-APT further combines clean-action $x$-prediction with asymmetric \emph{near--far} attention: executable near-term actions cannot attend to speculative far-future actions, while later actions build upon earlier ones.

To scale MM-ABC across embodiments, we construct a heterogeneous pretraining mixture containing \textbf{5,000+ hours of robot data}, spanning \textbf{12 datasets, 51 subsets, 400K+ episodes, and 17 embodiments}.
We further collect \textbf{MM-30}, a real-world mobile manipulation dataset with \textbf{30+ hours} of multi-view demonstrations on a HexFellow Trigger-A3 omnidirectional base with two AgileX PiPER-X 6-DoF arms, covering \textbf{40+ tasks} with broad object, background, and skill diversity.
A unified canonical action space and embodiment-aware masking preserve embodiment-specific control structure across heterogeneous sources.

We evaluate MM-ABC on \textbf{EBench, RoboCasa365, ManiSkill-HAB, LIBERO, and LIBERO-Plus}, together with five real-world mobile manipulation tasks (Section~\ref{sec:experiments}).
MM-ABC achieves 44.71\% success on EBench and a task-weighted average of 61.2\% on RoboCasa365, exceeding the strongest baselines in the respective comparisons by 3.30 and 7.0 percentage points.
It also achieves 99.1\% mean success on LIBERO, and the same LIBERO-trained policy reaches 82.8\% on LIBERO-Plus without further training, the highest among the compared methods, including those trained on perturbed demonstrations.
On five real-world tasks, MM-ABC achieves 83\% mean success, 12 points above $\pi_{0.5}$, supporting the use of a shared architecture across fixed-base and mobile settings.
These results suggest that scalable mobile manipulation requires more than adding mobility to a manipulation policy: the model must \emph{see} a changing workspace, \emph{imagine} how interaction will reshape it, and \emph{coordinate} arm and base through representations designed for collaboration.

Our contributions are summarized as follows:
\begin{itemize}
    \item We present \textbf{MM-ABC}, a large-scale foundation model for mobile manipulation built around Seeing, Coordinating, and Imagining Arm--Base Collaboration.

    \item We introduce sparse \textbf{DeepStack} visual conditioning and training-only \textbf{world imagination}, using future geometry as extra supervision of geometric intent so that perception and manipulation-intent prediction receive a denser learning signal without deployment-time overhead.

    \item We propose \textbf{MM-APT}, a structured dual-stream action transformer with clean-action $x$-prediction and near--far attention.
    A controlled synthetic study and a component ablation on RoboCasa365 both favor clean-action prediction over velocity prediction.

    \item We collect \textbf{MM-30}, a real-world mobile manipulation dataset covering \textbf{40+ tasks} and \textbf{30+ hours} of multi-view demonstrations with diverse objects, backgrounds, and everyday skills.

    \item We pretrain on \textbf{5,000+ hours, 400K+ episodes, 12 datasets, and 17 embodiments}, and evaluate mobile and fixed-base manipulation through simulation benchmarks, controlled ablations, and real-world deployment on five household, office, workcell, and laboratory tasks.
    MM-ABC delivers strong results throughout, with clear leads on the mobile manipulation benchmarks, strong fixed-base performance on LIBERO and LIBERO-Plus, and 83\% mean success in the real world.
\end{itemize}

% Requires natbib (\citep); bibliography: main.bib.
\section{Related Work}
\label{sec:related_work}

\subsection{Mobile Manipulation}
\label{sec:rw_mobile_manip}

Mobile manipulation couples base motion with object interaction, extending robot control beyond a fixed workspace.
Early model-based whole-body controllers~\citep{khatib1999} were followed by reinforcement and imitation learning approaches~\citep{hu2023causal,ehsani2024spoc,fu2024mobilealoha,jiang2025brs,chen2026mopa}, including whole-body coordination for dynamic grasping with legged manipulators~\citep{liang2026legged}.
Generalist policies now support discrete action decoding~\citep{brohan2023rt2,kim2024openvla}, diffusion~\citep{chi2023diffusionpolicy}, and flow matching~\citep{black2024pi0}, with mobile extensions addressing unfamiliar environments~\citep{pi2025pi05}, trajectory optimization~\citep{wu2025momanipvla}, and joint mobility--manipulation modeling~\citep{chen2026abotm05}.
Xiaomi-Robotics-1 scales UMI pretraining beyond 100K hours before aligning the policy with robot embodiments and instructions~\citep{team2026xiaomi}.
LingBot-VLA 2.0 combines broader pretraining with whole-body action interfaces and predictive semantic and geometric supervision~\citep{wu2026foundationapplication}.
These systems broaden the range of tasks and embodiments a single policy can support, while leaving coordination between heterogeneous control channels an important architectural concern.

Beyond scale, architecture determines how perceptual features and action streams interact.
VITRA concatenates an extracted VLM cognition feature with state and noisy action tokens in a DiT~\citep{li2025vitra}, while Qwen-VLA jointly processes VLM hidden states and noisy actions through self-attention~\citep{qwen2026vla}.
RLDX-1 extends this approach to modality-specific cognition, action, and optional physical-signal streams coupled through joint self-attention~\citep{kim2026rldx1}.
For coordination across body subsystems, AC-DiT conditions manipulation on a mobility prior~\citep{chen2025acdit}, InCoM and GeoHAT introduce structured arm--base coordination~\citep{liu2026incom,zhu2026geohat}, and MoPA aligns perception separately for each subsystem~\citep{chen2026mopa}.
DreamTrajectory guides whole-body actions with end-effector trajectories and refines candidates through a trajectory world model at test time~\citep{yang2026dreamtrajectory}.
For humanoids, $\omega$-0 combines future observation embeddings with controller-compatible action latents for concurrent locomotion and manipulation~\citep{li2026omega0}.
MM-ABC applies joint attention~\citep{esser2024mmdit} to specialized manipulation and mobility streams while keeping extracted VLM features as read-only context.
The action streams retain separate transformations and asymmetric temporal visibility; we study how the action prediction target affects denoising and task allocation in this two-stream setting.

\subsection{Representation Alignment and Predictive Supervision}
\label{sec:rw_rep_world}

Spatially grounded policies incorporate three-dimensional scene representations~\citep{shridhar2023peract,ke20243dda}, point-cloud conditioning~\citep{li2025pointvla,sun2025geovla}, and spatially informed VLA architectures~\citep{qu2025spatialvla,li2025spatial3d}.
Further work models temporal structure~\citep{zhang2025fourdvla,liu2026lift3dvla}, learns dynamic-aware 3D representations for scalable robot learning~\citep{liang2026afro}, or integrates geometry with semantic features~\citep{lin2025evo0,chen2026geoalign,liu2026vistavla,tu2026sgvla}.
The interface to the action expert also matters: policies use cross-attention~\citep{nvidia2025groot,shukor2025smolvla}, final-layer features~\citep{yang2026abotm0}, or predictive embeddings~\citep{miao2026jepavla}.
DeepStack, Qwen3-VL, and DeepVision-VLA further motivate conditioning across network depth~\citep{meng2024deepstack,bai2025qwen3vl,luo2026deepvision}.
MM-ABC uses sparse intermediate VLM features as read-only perceptual context.
This interface preserves multilevel visual information without requiring action tokens to modify the VLM token sequence.

Complementary approaches improve representations through teacher supervision, building on feature alignment and distillation~\citep{yu2025repa,shang2024theia}.
Spatial Forcing~\citep{li2026spatialforcing} and GLaD~\citep{guo2025glad} supervise internal policy features with geometry; ROCKET extends alignment across layers~\citep{sun2026rocket}, while VEGA targets the visual encoder~\citep{wang2026vega}.
Supervision ranges from object geometry~\citep{ding2026mindvla,liu2026sam3d}, depth~\citep{li2025qdepth}, and attended-region reconstruction~\citep{song2025reconvla} to spatial and semantic knowledge~\citep{shi2026think3d,tong2026xsvla,li2026lss}, latent actions~\citep{liu2026lara}, and temporal geometry~\citep{ding2026temporalforcing}.
These methods differ both in the information supplied by the teacher and in the part of the policy receiving supervision, making the alignment pathway an important design choice.

Predictive supervision extends these ideas to future images~\citep{wu2024gr1}, video embeddings~\citep{assran2025vjepa2}, and latent world representations~\citep{zheng2025flare,sun2026vlajepa,zhang2025dreamvla}, including geometric evolution~\citep{han2026gam,yang20264dwam}.
PHR-VLA, WAM4D, and MECo-WAM use removable prediction branches to retain training benefits without deploying the auxiliary predictor~\citep{soleymanzadeh2026phrvla,li2026wam4d,zhang2026mecowam}.
MM-ABC similarly predicts future VGGT-$\Omega$ features during training~\citep{wang2026vggtomega}.
Its future and action streams remain mutually masked, so geometric supervision shapes shared perceptual features without making future tokens policy inputs.
The resulting design complements current-scene conditioning with a predictive learning signal, while removing the teacher and future branch at deployment.

\subsection{Clean-Action Prediction and Cross-Stream Interaction}
\label{sec:rw_xpred}

The prediction target is a longstanding design choice in diffusion and flow models~\citep{salimans2022distillation,karras2022edm,lipman2023flowmatching}, including Transformer-based generators~\citep{peebles2023dit}.
In robotics, direct clean-action prediction already spans several settings.
DP3 uses sample prediction for high-dimensional action generation~\citep{ze2024dp3}, and RDT-1B trains a denoising network to regress clean action chunks for bimanual manipulation~\citep{liu2024rdt}.
ManiCM adopts action-sample prediction within consistency distillation for one-step control~\citep{gao2024manicm}; FA-RDP likewise uses action-space reparameterization in distillation for reactive contact-rich manipulation~\citep{zhuo2026fardp}.
More recently, ABot-M0 combines clean-action outputs with a velocity-space objective~\citep{yang2026abotm0}, while VGFM uses action-space predictions for intermediate value guidance in offline reinforcement learning~\citep{koirala2026vgfm}.

Related image-generation studies provide complementary perspectives: JiT motivates clean-data prediction through manifold structure and finite capacity~\citep{li2025jit}, Pixel MeanFlow separates clean outputs from velocity-based objectives~\citep{lu2026pixelmeanflow}, and MiniT2I explores a minimalist generation framework~\citep{wang2026minit2i}.
Across these settings, network outputs, training objectives, and sampling procedures are separate design choices: clean-action outputs can be trained with velocity-space losses, and consistency distillation differs from ordinary denoising or flow training.
Although endpoint and velocity predictions are algebraically convertible along an affine flow path, loss weighting and preconditioning can change their optimization behavior.
Controlled robot-policy studies reinforce the importance of separating these effects from architecture and iterative prediction~\citep{pan2025noising}.
Building on established clean-action prediction, MM-ABC studies its role in coupled arm--base streams: how the parameterization affects the noise retained in endpoint estimates, denoising accuracy across noise levels, and the arm--body task allocation realized by few-step sampling.
We evaluate these effects against an analytical Bayes-optimal denoiser and in full-policy ablations.

% ==============================================================================
\section{Understanding Clean-Action Prediction}
\label{sec:toy}

Before introducing MM-ABC, we examine whether a two-stream action network should predict clean trajectories or flow velocities.
Following JiT's distinction between prediction and loss spaces~\citep{li2025jit}, we compare both parameterizations under a common objective on a synthetic task whose action geometry and Bayes-optimal denoiser are known in closed form.
The two variants share architecture, conditioning, data, noise, and loss, and differ only in what the network emits.

\paragraph{Parameterizations.}
For a clean chunk $x$ and Gaussian noise $\epsilon\sim\mathcal N(0,I)$, flow time $t\in[0,1]$ defines
\begin{equation}
    z_t=(1-t)\epsilon+tx,
    \qquad v=x-\epsilon.
    \label{eq:flow}
\end{equation}
An $x$-head estimates $x$ directly, whereas a $v$-head estimates $v$.
With shared conditioning $c$, the two networks produce the following endpoint estimates of the clean chunk:
\begin{equation}
    \hat x^{(x)}=f_\theta(z_t,t,c),
    \qquad
    \hat x^{(v)}=z_t+d(t)g_\theta(z_t,t,c),
    \label{eq:toy_heads}
\end{equation}
where $d(t)=\max(1-t,0.05)$ stabilizes conversion near the clean endpoint.
Both minimize masked endpoint MSE with inverse-square weights $d(t)^{-2}$, normalized to unit minibatch mean, which equals velocity regression for the $v$-head wherever $d(t)=1-t$.
Both variants recover the sampling velocity as $(\hat x-z_t)/d(t)$.

\paragraph{Synthetic task.}
We generate paired manipulation and body chunks of shapes $16\times58$ and $16\times22$ online.
A condition $c_0\in\mathbb R^{12}$ maps to an eight-dimensional task vector $\tau$, and an equiprobable mode $m\in\{0,1\}$ sets the arm share $\alpha_m\in\{0.35,0.65\}$:
\begin{equation}
    \operatorname{vec}(x_s)
    =a_{e,s}Q_{e,s}
    [\rho_s(m)\tau;\,b_s(2m-1);\,u_s],
    \label{eq:toy_data}
\end{equation}
where $s\in\{M,B\}$, $\rho_M=\alpha_m$, $\rho_B=1-\alpha_m$, $b_M=0.12$, $b_B=1.60$, and $u_s\sim\mathcal N(0,I_4)$ is private variation.
Each $Q_{e,s}$ has 13 orthonormal, temporally smooth columns on the valid channels of embodiment profile $e$, one of six profiles, two of which have no body supervision.
Both networks receive $(z_{t,M},z_{t,B},t,c_0,e,m)$ with channel masks, so the arm--body allocation is given and the comparison isolates the prediction target.
A four-block, width-512 transformer uses separate stream transformations, masked cross-stream attention, and no raw-input skip inside either decoder.
Paired runs share initialization, batches, noise, and flow times, and train for 12,000 updates with batch size 256 and $t\sim\mathrm{Beta}(1.5,1)$; we report eight paired seeds, or sixteen seeds for the time-resolved metrics in panels (b) and (c).

\begin{figure*}[t]
    \centering
    \includegraphics[width=\linewidth]{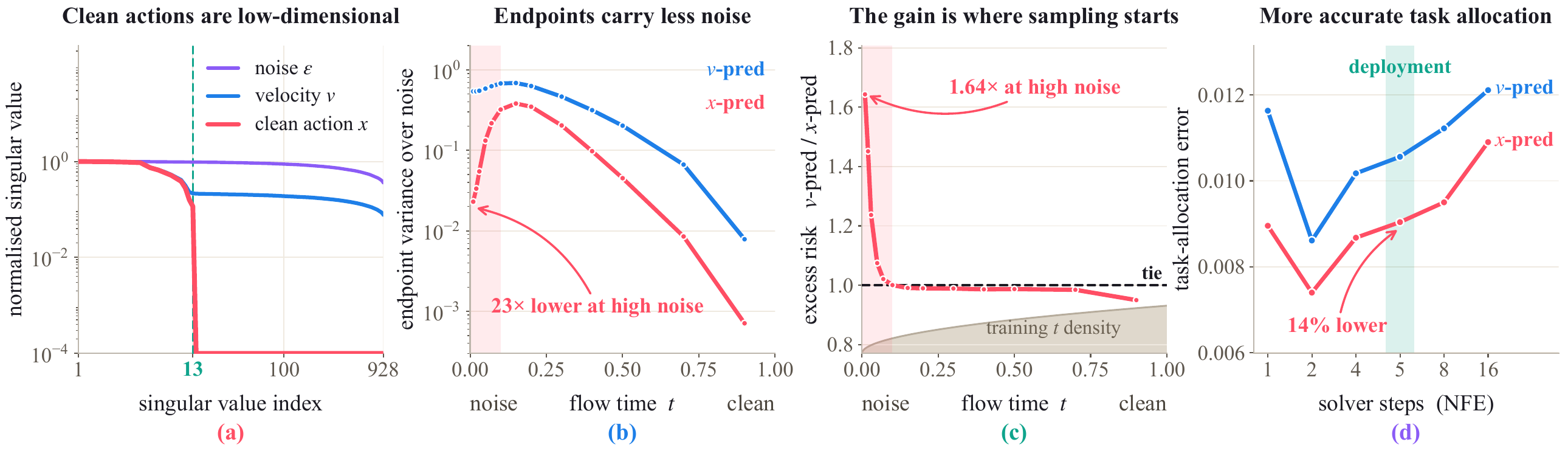}
    \vspace{-0.5em}
    \caption{\textbf{Clean-action versus velocity prediction under a common loss and a given arm--body allocation.}
    \textbf{(a)} Clean chunks occupy a prescribed 13-dimensional subspace, whereas velocity and noise span the ambient space.
    \textbf{(b)} Variance of the endpoint estimate across noise draws with the clean chunk fixed.
    \textbf{(c)} Excess risk over the Bayes-optimal denoiser, as a $v$/$x$ ratio.
    \textbf{(d)} Task-allocation error of sampled chunks against the number of Euler steps.}
    \label{fig:xpred_mechanism}
\end{figure*}

\paragraph{(a) Target geometry.}
For one embodiment, we stack flattened 928-dimensional manipulation chunks into matrices for $x$, $v$, and $\epsilon$, center them, and plot normalized singular values.
The clean spectrum terminates at rank 13, as prescribed by the generator, while velocity and noise retain variation across the ambient space.
A $v$-head must therefore reproduce noise directions that are absent from the clean-action subspace.

\paragraph{(b) Noise in the endpoint estimate.}
We compare both heads through their endpoint estimates $\hat x$, which live in the same space.
Holding a clean chunk fixed, we draw 24 independent noises, compute $\hat x$ for each, and report its variance normalized by the clean-chunk variance.
$x$-prediction yields lower variance at every tested flow time: $0.023$ versus $0.54$ at $t=0.01$, a $23\times$ reduction, and $0.35$ versus $0.63$ at $t=0.2$.
Ridge probes on the final manipulation-token features show the same pattern internally: the injected noise is recoverable with held-out $R^2$ of $0.62$ for $x$-prediction and $0.94$ for $v$-prediction, against $-0.09$ for both with permuted rows.
Because neither decoder has a raw-input skip, a $v$-head must carry this noise through its layers to form its endpoint estimate $z_t+d(t)g_\theta$ from the noisy input.

\paragraph{(c) Error beyond the Bayes-optimal denoiser.}
At fixed $t$, we compute endpoint MSE over valid entries and subtract the MSE of the Bayes-optimal conditional mean given the same inputs, available analytically in the known basis.
This excess risk measures approximation error beyond the uncertainty inherent in denoising.
The $v$/$x$ ratio reaches $1.64$ at $t=0.01$, so $x$-prediction is markedly more accurate in the high-noise regime where every sampling trajectory begins.

\paragraph{(d) Sampled trajectories.}
We sample from Gaussian noise with uniform Euler steps and read each generated chunk out in the known latent coordinates.
The task-allocation error is the distance between each stream's realized share of $\tau$ and the nearest valid share, averaged over the two streams; it counts coordination errors even when the chunk stays inside the action subspace.
$x$-prediction lowers this error at every tested budget from 1 to 16 steps, from $0.0106$ to $0.0090$ at five steps, the budget used by our policy.
It also yields lower off-subspace mass at every tested budget ($0.095$ versus $0.104$ at five steps).
Its sliced Wasserstein distance to the true latent distribution, which also reflects the private variation $u_s$, is lower at 8 and 16 steps ($0.075$ versus $0.076$ and $0.046$ versus $0.050$).

In this task, the allocation is supplied through $m$.
In MM-ABC, it is inferred from observations, the instruction, and robot state, and the two action streams exchange it through masked joint attention (\cref{sec:mmjit}).
The component ablation in \cref{sec:ablations} evaluates clean-action prediction in this learned setting, where it outperforms velocity prediction by 3.6 points on RoboCasa365 composite-seen tasks.

% ==============================================================================
\section{MM-ABC Architecture}
\label{sec:method}

\subsection{Overview}
\label{sec:overview}

At environment step $n$, MM-ABC receives current multi-view images $I_n$, a language instruction with embodiment and control metadata, and normalized robot state $s_n\in\mathbb R^{80}$ with validity mask $m_n^s\in\{0,1\}^{80}$.
It generates a 64-step normalized action chunk $X\in\mathbb R^{64\times80}$, split into manipulation $X_M\in\mathbb R^{64\times58}$ and body motion $X_B\in\mathbb R^{64\times22}$.
Embodiment-specific masks identify supervised entries; native control mappings are described in \cref{sec:data_engine}.
We distinguish environment step $n$ from flow time $t\in[0,1]$.

\begin{figure*}[t]
    \centering
    \includegraphics[width=\linewidth]{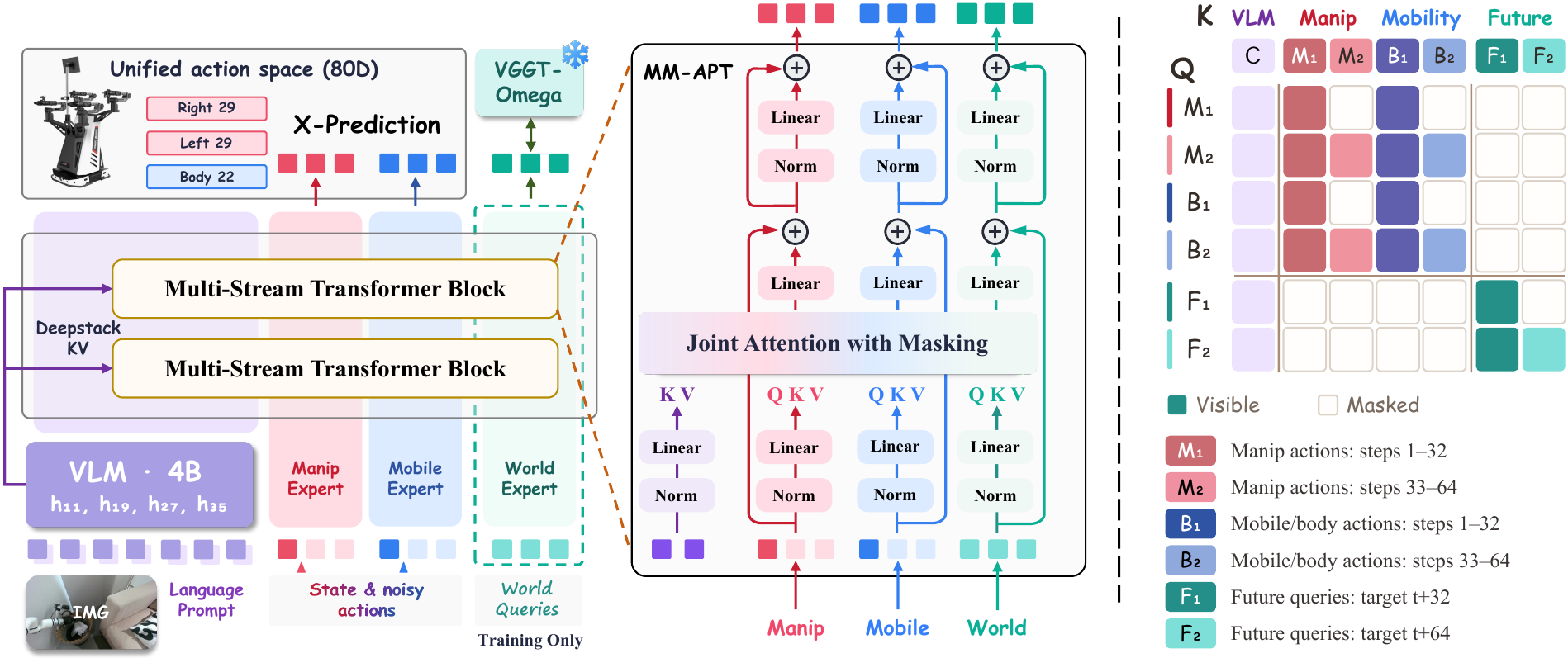}
    \vspace{-0.5em}
    \caption{\textbf{MM-ABC architecture.}
    \textbf{Left:} sparse VLM features condition MM-APT; a training-only future branch receives geometric supervision from frozen VGGT-Omega features.
    \textbf{Middle:} joint transformer blocks (MM-JiT in the diagram) combine stream-specific transformations with masked attention and read-only perceptual context.
    \textbf{Right:} action streams communicate within each segment, and far tokens additionally read near tokens. Future queries follow the same temporal ordering but remain isolated from actions.}
    \label{fig:pipeline}
\end{figure*}

As shown in \cref{fig:pipeline}, a Qwen3-VL-4B backbone encodes the current images and instruction.
The \textbf{Mobile Manipulation Action Prediction Transformer (MM-APT)} reads this context and jointly denoises two action streams using the clean-endpoint interface of \cref{sec:toy}.
A training-only future stream predicts later geometric features from the same context, while remaining isolated from action tokens.

\subsection{Sparse Multilevel Perceptual Context}
\label{sec:deepstack}

Mobile manipulation requires both recognizing the intended interaction and resolving local geometry as the viewpoint changes.
Conditioning only on the final VLM layer places both demands on a single representation optimized for the backbone's output tasks.
Features at intermediate depths offer additional access to visual information that may be less explicit in the final representation, motivating conditioning across depth~\citep{meng2024deepstack,luo2026deepvision}.
At the other extreme, separately attending to every VLM layer would introduce many feature interfaces for the action expert to reconcile during training.
We instead retain the final representation as a common semantic context and add a small number of intermediate features through gated residual updates.

We extract hidden states $H^l\in\mathbb R^{S\times2560}$ from zero-indexed VLM layers $l\in\{11,19,27,35\}$, where $S$ is the image--text sequence length.
Starting from $C_{-1}=H^{35}$, we inject intermediate features into the first three of the 16 expert blocks as gated residual updates:
\begin{equation}
    C_i = C_{i-1}+g_i\odot P_i\!\left(\operatorname{LN}(H^{l_i})\right),
    \qquad i=0,1,2.
    \label{eq:deepstack}
\end{equation}
Here $(l_0,l_1,l_2)=(11,19,27)$, $\operatorname{LN}$ denotes layer normalization, $P_i$ is a tokenwise linear map preserving width 2560, and $g_i\in\mathbb R^{2560}$ is a zero-initialized gate; $\odot$ denotes elementwise multiplication broadcast over tokens.
Later blocks retain $C_i=C_{i-1}$.
Each block projects its context into width-1024 keys and values, which action tokens read without updating the context sequence.
Gradients still reach the VLM, and context projections are reused across noise draws and sampling steps.
The zero-initialized gates make the initial interface identical to final-layer conditioning, allowing intermediate features to enter gradually as their gates are learned.
Sparse, cumulative injection limits changes in the context source across expert depth, and the read-only design holds perceptual evidence fixed throughout each denoising trajectory.
These choices provide a controlled route to multilevel information without requiring a separate connection to every backbone layer.

\subsection{Structured Two-Stream Action Generation}
\label{sec:mmjit}

Manipulation and body motion contribute differently to the same task: the body changes reachability and viewpoint, while the arm controls local interaction.
Separate transformations accommodate these different control semantics, but independent predictors would have to infer their partner's motion indirectly.
Joint attention lets each stream adjust its prediction using the other's evolving action representation, while both remain grounded in the same observation and full robot state.

\paragraph{Token construction.}
MM-APT uses 16 transformer blocks of width 1024, with 16 attention heads and feed-forward width 4096.
Separate stream encoders project the noisy actions $Z_{t,s}$, concatenate a broadcast sinusoidal time embedding, and fuse them through a multilayer perceptron, for $s\in\{M,B\}$.
Each stream prepends a state token encoded from the full $[s_n;m_n^s]\in\mathbb R^{160}$, yielding 65 tokens.
Learned position and segment embeddings distinguish action steps and divide the chunk into near (state token and first 32 actions) and far (remaining 32 actions) segments.

\paragraph{Structured communication.}
Streams retain separate normalization, attention projections, and SwiGLU feed-forward layers, but their tokens participate in joint masked attention with read-only perceptual context.
Manipulation and body tokens communicate bidirectionally within each segment.
Far queries additionally read near keys, whereas near queries cannot read far keys, allowing the tail to build on the immediate plan without influencing it.
This asymmetry reflects the unequal roles of the two horizons: the prefix must support the next physical interaction, while the tail anticipates states that will be observed again before execution.
For example, an immediate reach should be grounded in the current object and base configuration, while a later repositioning can be refined after new observations arrive.
The distinction between immediate execution and longer-horizon organization also appears in accounts of hierarchical motor control~\citep{merel2019hierarchical}.
Both horizons supervise the shared parameters under this near-to-far attention pattern.

Future queries obey the same temporal ordering but cannot exchange information with either action stream.
All streams read valid context tokens; padding and inactive body streams are masked.
After the final block, separate normalized multilayer decoders discard the state tokens and predict clean chunks $\hat X_M$ and $\hat X_B$, without a direct noisy-input skip.

\subsection{Future Geometry as Auxiliary Supervision}
\label{sec:future}

Current-scene understanding alone does not specify which spatial relations matter for the next interaction.
A robot approaching a handle, for example, needs features informative about how the handle and end effector will come into alignment, as well as the handle's present appearance.
This motivates training the VLM representation to support anticipation of task-relevant geometry.
Predicting future geometric features provides spatially distributed supervision beyond the action vector: successful prediction requires preserving information about the scene that helps explain its subsequent configuration.
Using features from a geometry-oriented teacher~\citep{wang2026vggtomega} directs this objective toward spatial structure, while future rather than current targets encourage the representation to capture how that structure evolves during the demonstrated task.

The training-only World Expert predicts geometric features at environment steps $n+32$ and $n+64$ from current perceptual context.
Each horizon has 192 learned width-1024 queries, corresponding to an $8\times8$ grid for each of up to three camera views.
These queries form a separate stream in the joint transformer and receive neither action tokens nor future images.

A frozen VGGT-Omega teacher processes recorded future images separately at each horizon and pools its features to the same spatial grids.
A layer-normalized linear decoder maps future-query outputs to 2048-dimensional predictions $\hat F_{brj}$ matching teacher targets $F^*_{brj}$, where $b$, $r$, and $j$ index examples, horizons, and view--spatial tokens.
The alignment objective is
\begin{equation}
    \mathcal L_{\mathrm{future}}
    =\frac{\sum_{b,r,j}q_{brj}
        \left[1-\operatorname{cos}_{\delta}(\hat F_{brj},F^*_{brj})\right]}
        {\max(1,\sum_{b,r,j}q_{brj})}.
    \label{eq:future_loss}
\end{equation}
Here $q_{brj}\in\{0,1\}$ masks unavailable future frames, views, and invalid teacher vectors; teacher targets receive no gradients.
The stabilized cosine is $\operatorname{cos}_{\delta}(u,v)=u^{\mathsf T}v/[\sqrt{\|u\|_2^2+\delta}\sqrt{\|v\|_2^2+\delta}]$ for a small $\delta>0$.
This auxiliary predictor supervises the shared perceptual representation to anticipate demonstrated geometry.
Because the future stream reads the same VLM features as the action streams, its loss trains the backbone to make this predictive geometric information available to the policy.
Future observations define training targets, while action generation reads the current perceptual context under the mutual attention mask.
Both the future stream and teacher are removed at inference.

\subsection{Training Objective}
\label{sec:xpred}

We construct $Z_t=(1-t)E+tX$ using independent Gaussian noise $E$ and $t\sim\mathrm{Beta}(1.5,1)$ clipped at $0.999$, and predict $\hat X$ from current context and state.
For a supervised minibatch of size $B$, let $M_{b,s}$ mask valid action entries and $\mathcal S_b$ contain the active streams of example $b$.
Using $d(t)$ from \cref{sec:toy}, we define the normalized weight $\bar w_b=d(t_b)^{-2}/[B^{-1}\sum_{b'=1}^{B}d(t_{b'})^{-2}]$.
The masked action loss is
\begin{equation}
    \mathcal L_{\mathrm{action}}
    =\frac{1}{B}\sum_{b=1}^{B}
       \frac{\bar w_b}{|\mathcal S_b|}
       \sum_{s\in\mathcal S_b}
       \frac{\|M_{b,s}\odot(\hat X_{b,s}-X_{b,s})\|_F^2}
            {\|M_{b,s}\|_1}.
    \label{eq:action_loss}
\end{equation}
The squared Frobenius norm $\|\cdot\|_F^2$ sums squared errors, and $\|M_{b,s}\|_1$ counts valid entries, giving active streams equal weight regardless of their supervised dimensions.
We average this loss over four independent noise/time draws sharing one context computation and evaluate future supervision once:
\begin{equation}
    \mathcal L=1.0\,\mathcal L_{\mathrm{action}}
        +0.05\,\mathcal L_{\mathrm{future}}.
    \label{eq:total_loss}
\end{equation}
At inference, the action streams generate chunks from Gaussian noise with five Euler steps, using the endpoint-to-velocity conversion in \cref{sec:toy}; the robot executes a prefix before replanning.

% ==============================================================================
% 5. Multi-Embodiment Pretraining Data Engine
% ==============================================================================

\section{Multi-Embodiment Pretraining Data Engine}
\label{sec:data_engine}

Robot datasets differ in control semantics, coordinate conventions, embodiment, and
sampling frequency. We construct MM-ABC's pretraining corpus by auditing each source,
standardizing its state and action representations, and retaining temporally contiguous
valid trajectories in a shared masked interface. A balanced sampling scheme controls
the contribution of each source to training.

After processing, our corpus contains \mmabcemph{5,166.2 hours} and \mmabcemph{400K+ episodes}
of demonstrations from \mmabcemph{12 datasets}, \mmabcemph{51 subsets}, and
\mmabcemph{17 embodiments}, together with \mmabcemph{65K+ unique natural-language instructions}.
Real-world and simulated demonstrations account for 80\% and 20\% of the total duration,
respectively. Platforms capable of mobile manipulation contribute approximately 61\% of the
cleaned hours. Within this group, 1,371.7 hours (26.6\% of the full corpus) both retain base-control
channels and exhibit nontrivial base motion.
Our self-collected mobile manipulation set contains \textbf{30+ hours} of demonstrations
across \textbf{40+ tasks}, with all cleaned demonstrations included in pretraining
(\cref{sec:selfcollect}).

% ------------------------------------------------------------------------------
\subsection{Corpus Composition and Training Mixture}
\label{sec:data_composition}

\begin{figure*}[t]
    \centering
    \includegraphics[width=\linewidth]{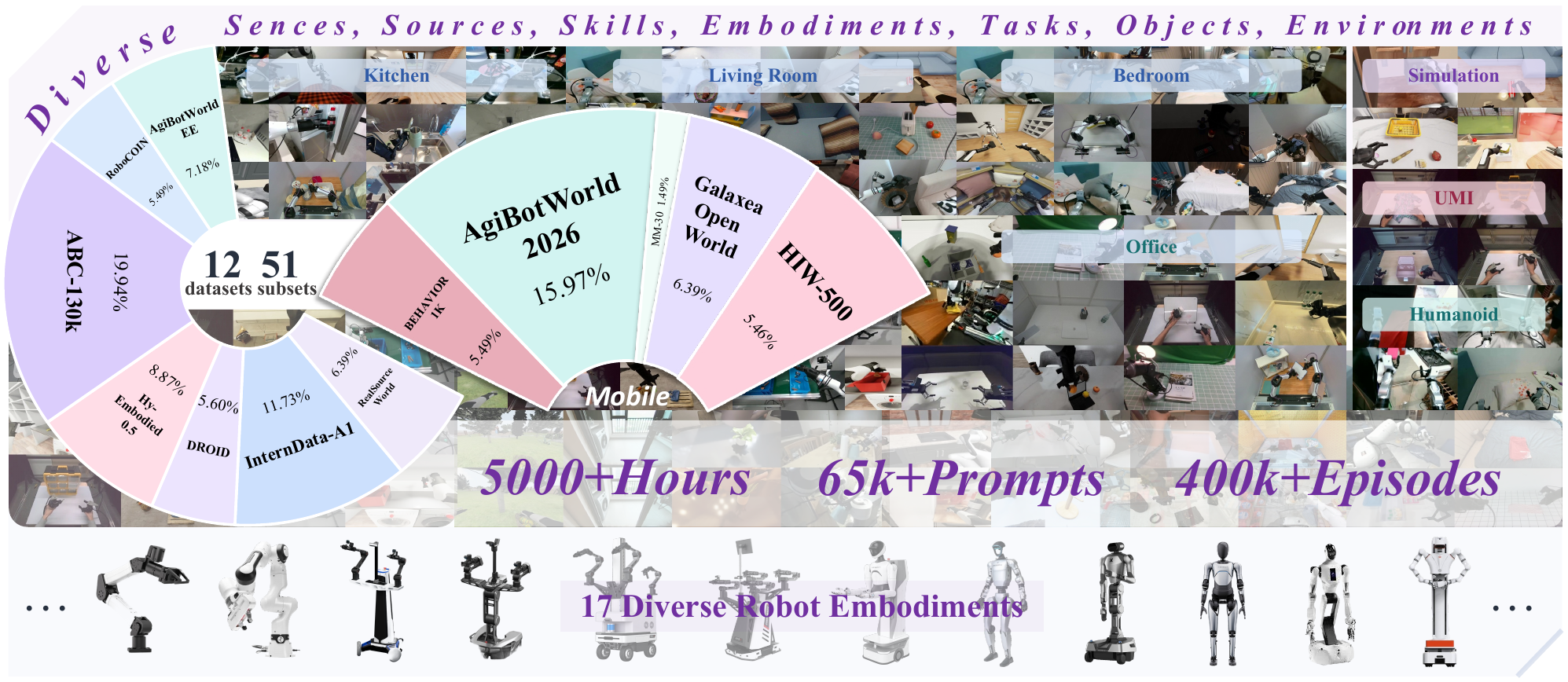}
    \vspace{-0.4em}
    \caption{
        \textbf{Overview of the MM-ABC multi-embodiment pretraining corpus.}
        The corpus contains 5,166.2 cleaned hours from 12 datasets and 51 subsets,
        spanning 17 embodiments. Sector areas show training sampling probabilities
        aggregated by dataset; the surrounding panels illustrate environments and embodiments.
    }
    \label{fig:data_pretrain}
\end{figure*}

\begin{figure*}[t]
    \centering
    \includegraphics[width=\linewidth]{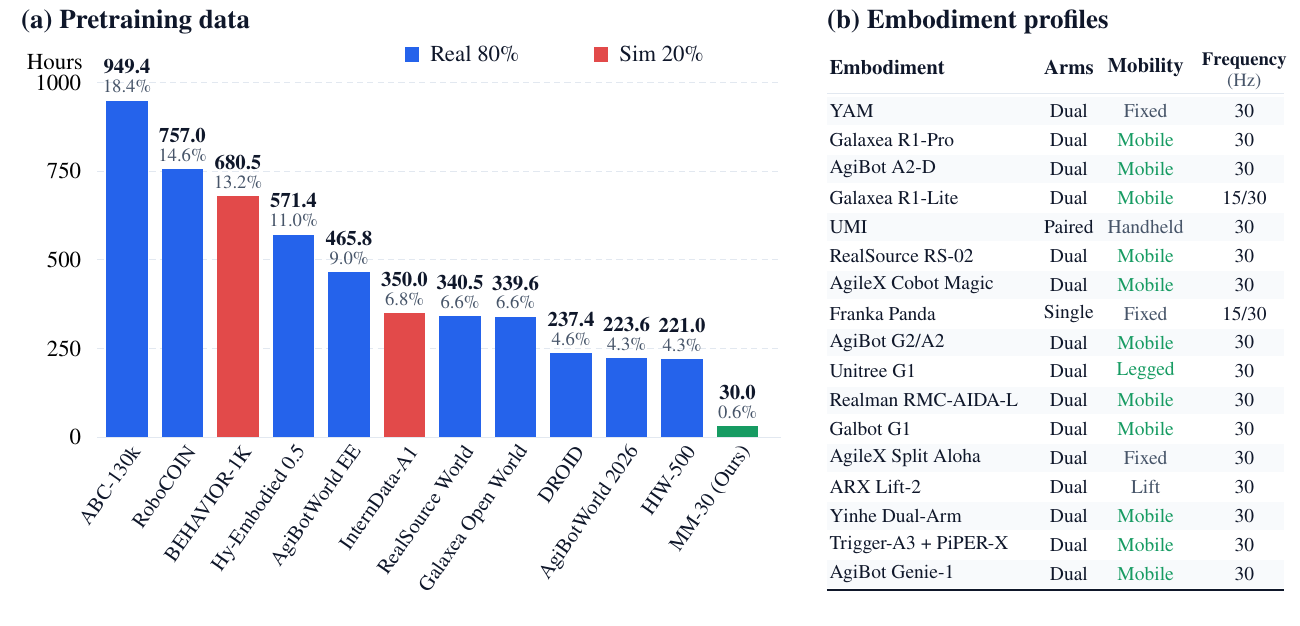}
    \vspace{-0.4em}
    \caption{\textbf{Pretraining data and embodiment composition.}
    \textbf{(a)} Dataset hours and duration shares.
    \textbf{(b)} Arm configuration, mobility, and control frequency.}
    \label{fig:corpus_summary}
\end{figure*}

Figure~\ref{fig:data_pretrain} summarizes the corpus and its \emph{training sampling
probabilities}; Figure~\ref{fig:corpus_summary}(a) reports the duration contributed by each
dataset. These quantities differ because training uses a rebalanced mixture.
Some sources contribute 700--900 cleaned hours or more, while smaller sources add
embodiments and control configurations with limited representation in the corpus.

\paragraph{Data sources.}
The corpus combines eleven public datasets with our MM-30 collection.
BEHAVIOR-1K and InternData-A1 are simulated; the other sources are recorded on physical
robots or, for Hy-Embodied 0.5, with a handheld UMI device.
\begin{itemize}
    \item \textbf{ABC-130k}~\citep{allshire2026abc}: a bimanual teleoperation dataset of
    134,806 episodes and 3,553 hours, collected on low-cost stations with two 6-DoF YAM arms.
    Its 195 tasks cover pick-and-place, folding, handover, insertion, tool use, and assembly.
    \item \textbf{RoboCOIN}~\citep{wu2025robocoin}: over 180K teleoperated bimanual
    demonstrations from 15 robotic platforms, spanning 421 tasks in 16 residential, commercial,
    and working scenarios, with hierarchical annotations ranging from trajectory-level concepts
    to frame-level kinematics.
    \item \textbf{BEHAVIOR-1K}~\citep{li2024behavior1k}: a simulation benchmark of 1,000 everyday
    household activities built on OmniGibson. Its 2026 challenge release provides 20,000
    teleoperated demonstrations of 100 long-horizon tasks on the wheeled bimanual Galaxea R1-Pro.
    \item \textbf{Hy-Embodied 0.5}~\citep{zhang2026hyvla}: bimanual demonstrations collected with
    a fingertip UMI device tracked by optical motion capture and recorded from head and wrist
    views. The full corpus exceeds 10,000 hours, and the public release provides 2,163 hours
    over 70+ tasks.
    \item \textbf{AgiBotWorld EE}~\citep{bu2025agibotworld}: AgiBot World contains 1,001,552
    trajectories and 2,976.4 hours over 217 tasks, 87 skills, and 106 scenes, collected by more
    than 100 AgiBot robots in five deployment domains.
    \item \textbf{InternData-A1}~\citep{tian2026interndata}: synthetic data from a compositional
    simulation pipeline, with over 630K trajectories and 7,433 hours across 70 tasks, 18 skills,
    and 227 scenes, covering rigid, articulated, deformable, and fluid objects on Franka Panda,
    AgileX Split Aloha, ARX Lift-2, and AgiBot Genie-1.
    \item \textbf{RealSource World}~\citep{realsource2025world}: 11,428 episodes of long-horizon
    manipulation over 35 tasks on the RS-02 dual-arm humanoid, recorded in kitchens, conference
    rooms, convenience stores, homes, and industrial settings with atomic-skill segments and
    per-episode quality assessments.
    \item \textbf{Galaxea Open-World}~\citep{jiang2025galaxea}: 500+ hours of real-world mobile
    manipulation over 150+ tasks in 50 scenes, collected on a single embodiment, the Galaxea
    R1-Lite with two 6-DoF arms, a 3-DoF torso, and an omnidirectional base, and annotated with
    bilingual subtask labels.
    \item \textbf{DROID}~\citep{khazatsky2024droid}: in-the-wild single-arm Franka Panda
    demonstrations comprising 76K trajectories and 350 hours across 564 scenes and 86 tasks.
    \item \textbf{AgiBotWorld 2026}~\citep{agibotworld2026}: real-world data collected on the
    AgiBot G2 platform in a free-form mode, covering commercial spaces such as retail stores as
    well as home scenarios.
    \item \textbf{HIW-500}~\citep{hiw500_2026}: 500+ hours and 23K+ episodes of whole-body
    teleoperation of Unitree G1 humanoids in 12 real homes in Southeast Asia, covering 10+
    household tasks with 161 subtask labels.
    \item \textbf{MM-30 (ours)}: 30+ hours of coordinated base and dual-arm demonstrations
    over 40+ tasks on our mobile manipulation platform (\cref{sec:selfcollect}).
\end{itemize}

\paragraph{Balanced sampling.}
Sampling weights are assigned to source-specific training profiles according to their
cleaned duration. For a profile $i$ containing $h_i$ hours, we apply a sublinear duration
weighting, followed by a cap on each profile's normalized sampling probability:
\begin{equation}
    \tilde{p}_i = h_i^{0.4},
    \qquad
    p_i = \operatorname{CapNorm}\!\left(\tilde{p}_i; 0.2\right).
    \label{eq:data_mixture}
\end{equation}
$\operatorname{CapNorm}$ normalizes the weights, limits each profile's probability to
0.2, and redistributes excess probability mass among uncapped profiles.
This reduces the concentration of sampling probability in the largest profiles.
Figure~\ref{fig:data_pretrain} aggregates these probabilities across profiles belonging
to the same dataset.

% ------------------------------------------------------------------------------
\subsection{Self-Collected Mobile Manipulation Dataset}
\label{sec:selfcollect}

\begin{figure*}[t]
    \centering
    \includegraphics[width=\linewidth]{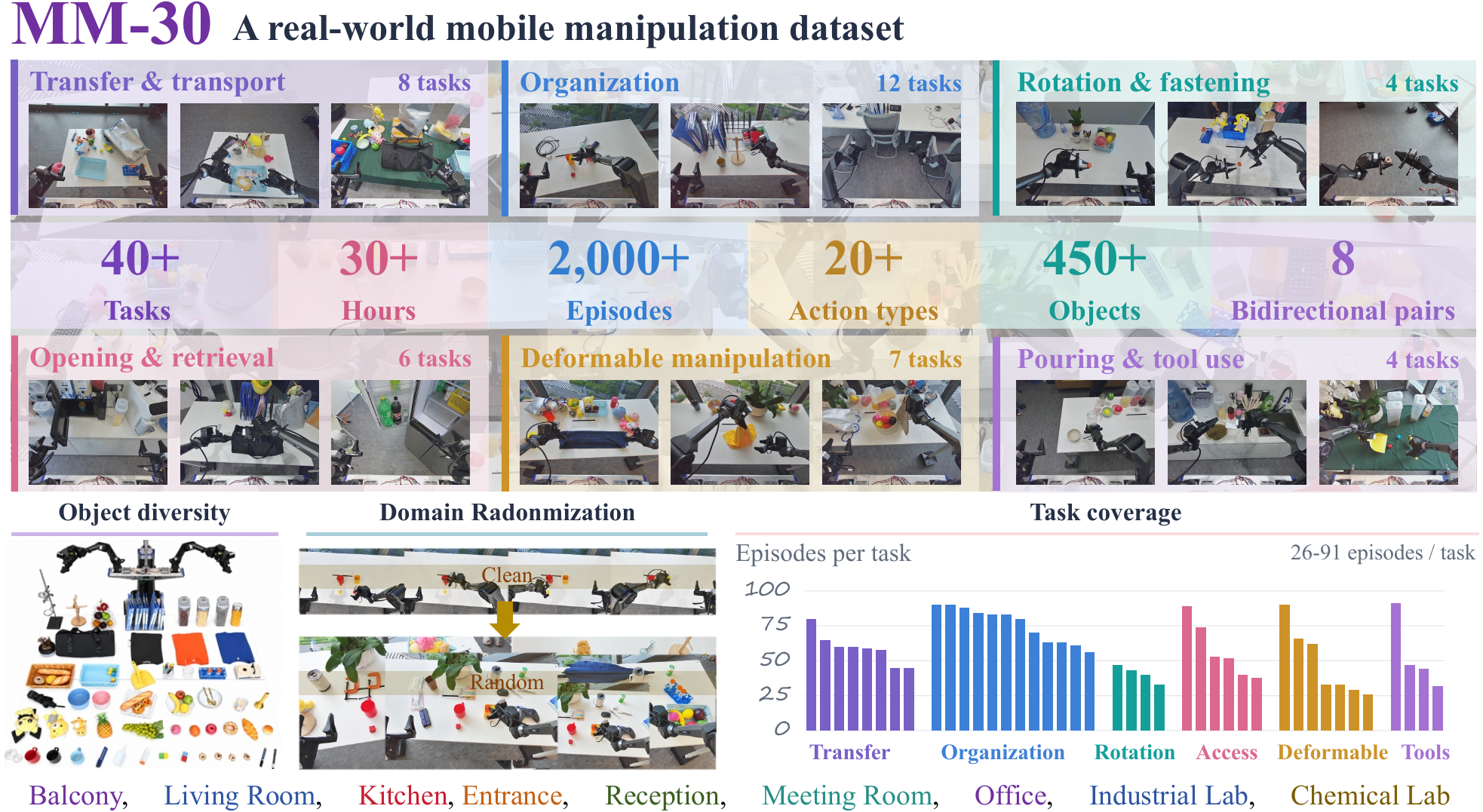}
    \vspace{-0.4em}
    \caption{\textbf{MM-30: self-collected mobile manipulation dataset.}
    Representative tasks, object and scene diversity, and episode counts per task.}
    \label{fig:mm30}
\end{figure*}

To complement public datasets, we collect \textbf{MM-30}, a real-world mobile
manipulation dataset with coordinated base and dual-arm motion.
Figure~\ref{fig:mm30} summarizes its tasks, objects, and scenes.

\paragraph{Hardware platform.}
The platform comprises a \textbf{HexFellow Trigger-A3} omnidirectional mobile base
and two \textbf{AgileX PiPER-X} 6-DoF manipulators.
The base repositions the robot in the horizontal plane, and the two arms perform object
interactions, including bimanual handovers.
Their active control channels map to the base and left- and right-arm slots of
the 80D interface (Figure~\ref{fig:corpus_summary}(b)).

\paragraph{Task coverage and diversity.}
We collect \textbf{30+ hours} of teleoperated demonstrations spanning \textbf{40+}
mobile manipulation tasks.
Multi-view demonstrations benefit manipulation learning beyond viewpoint
generalization~\citep{cai2026multiview}; we therefore record each task from multiple
viewpoints and vary initial configurations, object layouts, distractors, and scene
appearance to broaden visual and spatial coverage.
The demonstrations include picking, placing, opening, closing, pouring, wiping,
fetching, and short-horizon rearrangement, with base motion adjusting the
reachable workspace during manipulation.

\paragraph{Role in pretraining.}
After the semantic audit, coordinate alignment, and quality filtering described in
\cref{sec:data_cleaning}, all cleaned demonstrations from MM-30 are included in
MM-ABC pretraining (Figure~\ref{fig:corpus_summary}(a)).

% ------------------------------------------------------------------------------
\subsection{Embodiment Diversity and Unified Control Interface}
\label{sec:embodiment_diversity}

The corpus spans 17 embodiments, including single- and dual-arm systems,
fixed-base manipulators, wheeled mobile manipulators, a full-body humanoid, lift-equipped
platforms, and handheld UMI demonstrations. The number of valid stored dimensions ranges
from 10 to 43, and control frequencies range from 15 to 30\,Hz.
Figure~\ref{fig:corpus_summary}(b) summarizes their arm configurations, mobility,
and control frequencies.

For joint training across embodiments with different sensors and control channels, MM-ABC maps their state and action fields into a
fixed \mmabcemph{80D canonical state/action interface}:
\begin{equation}
    \mathbf{a}_t =
    \mathbf{a}^{L}_t \oplus
    \mathbf{a}^{R}_t \oplus
    \mathbf{a}^{B}_t
    \in \mathbb{R}^{29+29+22},
    \qquad
    \mathbf{m}_t \in \{0,1\}^{80},
    \label{eq:canonical_action}
\end{equation}
Here $\oplus$ denotes concatenation and $\mathbf{m}_t$ is an embodiment-specific validity mask.
The left and right 29D manipulation blocks reserve slots for arm joints, 3D end-effector
(EEF) position, 6D rotation representations, gripper state, and optional hand joints.
The 22D body block reserves slots
for mobile-base motion, torso, lift, head, and auxiliary controls.
Only channels supported by the verified source schema are populated.
Unavailable channels are set to zero and masked out of the action loss.

State and action use the same canonical field layout in the processed corpus.
When a source stores both joint and EEF representations, both can occupy the 80D vector;
the source's control mode determines which representation provides action supervision.
They are not treated as simultaneous, independent action targets.

% ------------------------------------------------------------------------------
\subsection{Data Engine: Semantic Audit, Cleaning, and Canonicalization}
\label{sec:data_cleaning}

\begin{figure*}[t]
    \centering
    \includegraphics[width=\linewidth]{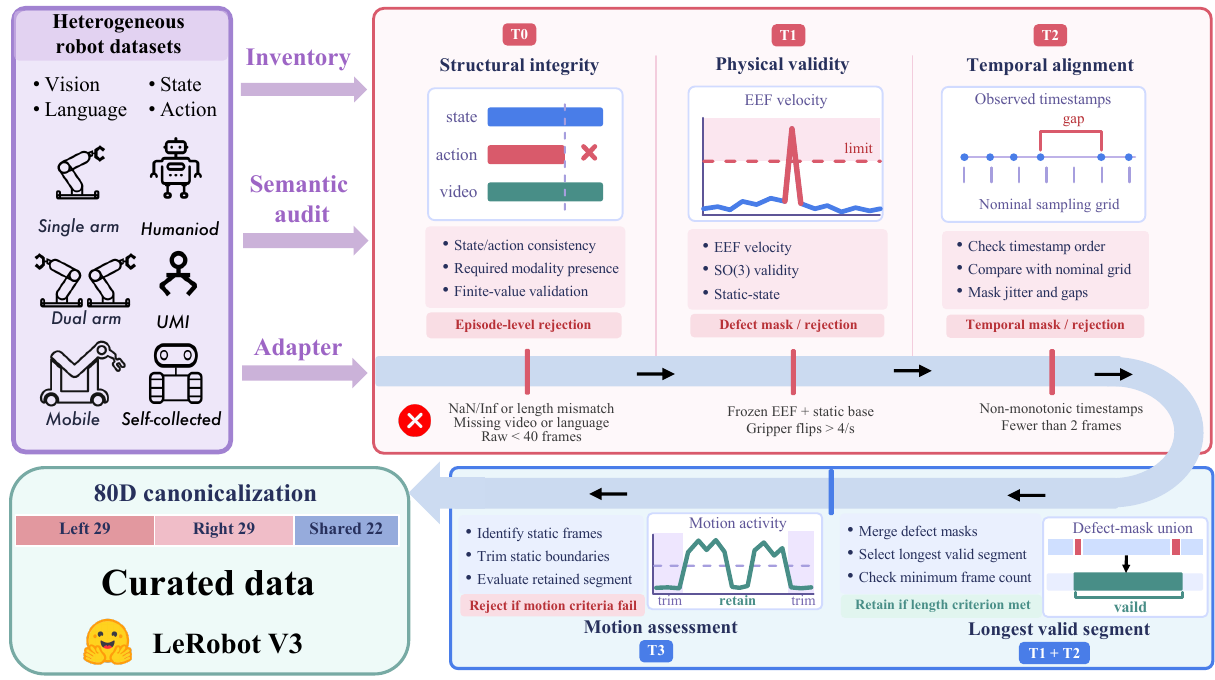}
    \vspace{-0.4em}
    \caption{
        \textbf{MM-ABC data engine.}
        Sources undergo semantic auditing, conversion with dataset-specific adapters,
        and four stages of quality filtering. Retained contiguous segments are encoded
        in the masked 80D interface, independently validated, and incorporated into the
        training mixture.
    }
    \label{fig:data_engine_pipeline}
\end{figure*}

The pipeline in Figure~\ref{fig:data_engine_pipeline} standardizes field semantics before
applying common quality checks. State and action tensors can encode joint targets,
Cartesian poses, deltas, velocities, or delayed controller commands.
Sources also differ in parent coordinate frames, tool center points (TCPs), rotation
conventions, and gripper conventions, requiring explicit interpretation before conversion.

\paragraph{Source audit and field semantics.}
For each source, we fix its revision and record its schema, camera streams, control
frequency, and robot metadata. A dataset-specific \emph{frame contract} specifies
the physical meaning of each state/action field: units, action type, parent frame, TCP,
rotation convention, gripper direction, and temporal semantics.
Dataset-specific adapters then convert units, unwrap Euler angles where needed,
reconstruct rotations in $\mathrm{SO}(3)$, and align coordinate frames and TCPs.
Explicit field semantics distinguish velocity commands from pose targets and identify
actions expressed relative to different tool centers, parent frames, or rotation conventions.

\paragraph{Staged quality filtering.}
Filtering proceeds from inexpensive structural checks to more expensive motion analysis.
\textbf{Tier 0} checks structural integrity, rejecting trajectories with insufficient length,
state/action length mismatches, non-finite values, missing required videos, or missing
language annotations.
\textbf{Tier 1} checks physical validity, including implausible EEF speed, workspace
violations, invalid rotations, frozen signals, and excessive gripper switching.
\textbf{Tier 2} checks timestamp monotonicity and flags temporal jitter and large gaps.
We merge the Tier~1--2 defect masks and retain the \emph{longest contiguous valid segment}.
Finally, \textbf{Tier 3} trims stationary prefixes and suffixes and rejects trajectories
dominated by stationary behavior or negligible displacement of the end effector and base.

Filtering preserves temporal continuity: invalid transitions delimit candidate segments,
and \emph{interior frames are never removed from a retained segment}.
This avoids introducing unrecorded time jumps into action chunks.
A two-pass implementation first records all retention, rejection, and trimming decisions
in a manifest, then writes the retained segments in the canonical format.

\paragraph{Independent validation.}
A separate validator checks the serialized data for schema consistency, finite values,
timestamp continuity, video seeking and decoding, rotation orthogonality and round-trip
consistency, compliance with the state/action frame contracts, and correct validity masks
and normalization statistics.
Corpus statistics and normalization parameters are recomputed from the cleaned data after
serialization.

% ==============================================================================
% 6. Experiments
% ==============================================================================
\section{Experiments}
\label{sec:experiments}

We organize our evaluation around three questions.
First, how does MM-ABC compare with strong VLA and WAM baselines on mobile and fixed-base manipulation, including under distribution shifts?
Second, how do sparse multilevel perceptual conditioning, clean-action prediction, and future geometric supervision contribute to performance?
Third, how effectively does MM-ABC coordinate base motion and object interaction on a physical robot?
To address these questions, we evaluate MM-ABC on EBench, RoboCasa365, ManiSkill-HAB, LIBERO, and LIBERO-Plus, conduct controlled component ablations, and compare policies on five real-world mobile manipulation tasks. We also integrated our implementation into XPolicyLab~\cite{community2026xpolicylab}.

% ------------------------------------------------------------------------------
\subsection{Simulation Experiments}
\label{sec:sim}

\begin{figure}[t]
    \centering
    \includegraphics[width=\linewidth]{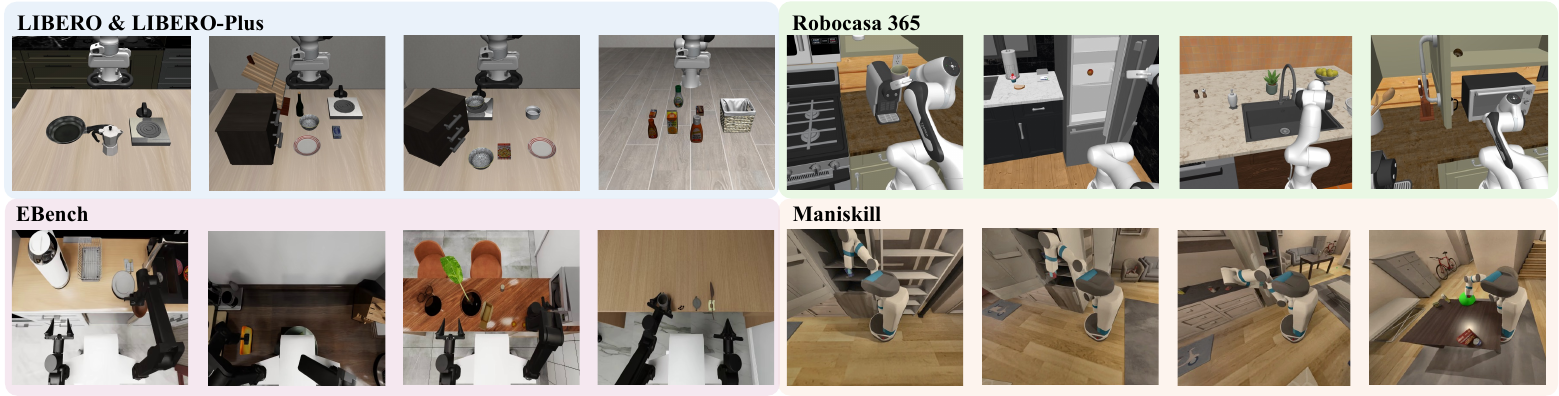}
    \vspace{-0.3em}
    \caption{\textbf{Simulation benchmarks.}
    Representative task scenes from LIBERO/LIBERO-Plus, RoboCasa365, EBench,
    and ManiSkill-HAB, covering fixed-base and mobile manipulation.}
    \label{fig:sim_benchmarks}
\end{figure}

\paragraph{Benchmarks and protocol.}
Figure~\ref{fig:sim_benchmarks} illustrates representative tasks from the simulation benchmarks.
We follow the official task splits and success criteria of each benchmark.
Starting from the pretrained MM-ABC weights, we adapt separate policies using the demonstrations available for each benchmark or task suite.
The comparisons include visuomotor policies, vision--language--action (VLA) models, and world-action models (WAMs).
Observation modalities and training budgets are given with each benchmark.
The benchmark comparisons assess the full policy, while \cref{sec:ablations} examines individual components under a common from-scratch training protocol on the RoboCasa365 composite-seen tasks.

\textbf{EBench.}
EBench comprises 26 indoor tasks spanning mobile pick-and-place, long-horizon mobile manipulation, and dexterous tabletop manipulation~\citep{gao2026ebench}.
Its tasks vary in scene, skill, horizon, precision, and operating mode, assessing both workspace repositioning and precise object interaction.
A single checkpoint is evaluated on the held-out test split following the official evaluation protocol, using task success and a stage-wise progress score.
We post-train MM-ABC for 100k steps with a batch size of 512.

\begin{figure}[t]
    \centering
    \includegraphics[width=\linewidth]{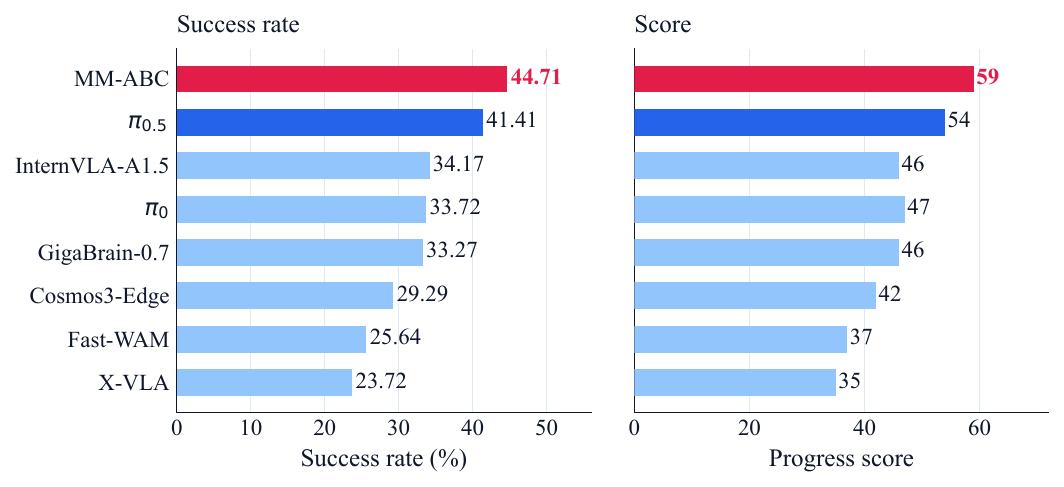}
    \vspace{-0.3em}
    \caption{
        \textbf{EBench evaluation.}
        Success rate (left) and stage-wise progress score (right) for MM-ABC and selected baselines~\citep{pi2025pi05,black2024pi0,internvla2026a15,gigabrain2026,nvidia2026cosmos3,yuan2026fastwam,zheng2025xvla}.
        Higher is better.
    }
    \label{fig:ebench}
\end{figure}

As shown in \cref{fig:ebench}, MM-ABC achieves a success rate of \textbf{44.71\%} and a progress score of \textbf{59}.
The strongest baseline, $\pi_{0.5}$, achieves 41.41\% and 54, respectively.
The corresponding improvements are 3.30 percentage points in success rate and 5 points in progress score, indicating gains in both task completion and intermediate progress.
This joint improvement is consistent with MM-APT's design, which conditions manipulation and body actions on shared multilevel context while accommodating both mobile and fixed-base control.

\textbf{RoboCasa365.}
The RoboCasa365 target evaluation comprises 50 household tasks in held-out kitchens: 18 atomic-seen, 16 composite-seen, and 16 composite-unseen tasks~\citep{nasiriany2026robocasa365}.
Atomic tasks emphasize individual interactions with objects and articulated fixtures, whereas composite tasks combine successive interactions with navigation.
The seen/unseen labels refer to task inclusion in RoboCasa365's predefined pretraining set.
We use the full set of target demonstrations for post-training, with 120k steps and a batch size of 512.

\Cref{tab:robocasa365} reports success rates for the three splits and their task-weighted average.
MM-ABC achieves 78.4\%, 52.7\%, and 50.3\%, respectively, with an overall average of 61.2\%.
These results exceed ABot-M0.5 by 7.8, 8.4, and 4.7 percentage points on the individual splits and by 7.0 points overall.
The improvements extend across atomic and composite tasks, with the largest margin on the composite-seen split.
Composite tasks chain several interactions with navigation and require the base and arms to act in concert as reachability and viewpoint change.
Across all compared methods, success rates are lower on composite tasks than on atomic tasks, which confirms composite tasks as the harder regime.

\begin{table}[t]
    \centering
    \small
    \setlength{\tabcolsep}{6pt}
    \caption{
        \textbf{RoboCasa365 target evaluation with full demonstrations}.
        Success rate (\%).
        S/U denote seen/unseen tasks; the average is weighted by the split sizes (18/16/16).
        Best: shaded; second best: bold.
    }
    \label{tab:robocasa365}
    \begin{tabular}{lcccc}
        \toprule
        \mmabctablehead Method & Atomic-S & Composite-S & Composite-U & Average \\
        \midrule
        GR00T-N1.5~\citep{nvidia2025groot} & 60.6 & 35.0 & 33.3 & 43.7 \\
        Fast-WAM~\citep{yuan2026fastwam} & 59.1 & 36.4 & 33.2 & 43.5 \\
        LingBot-VA~\citep{li2026lingbotva} & 63.5 & 37.3 & 32.1 & 45.1 \\
        ABot-M0.5~\citep{chen2026abotm05} & \secondbest{70.6} & \secondbest{44.3} & \secondbest{45.6} & \secondbest{54.2} \\
        \midrule
        \rowcolor{mmabcRow}
        \textbf{MM-ABC (Ours)} & \best{78.4} & \best{52.7} & \best{50.3} & \best{61.2} \\
        \bottomrule
    \end{tabular}
\end{table}

\textbf{ManiSkill-HAB.}
ManiSkill-HAB evaluates low-level mobile manipulation with a Fetch robot, including physically simulated grasping and interaction with articulated objects~\citep{shukla2025maniskillhab}.
SetTable requires retrieving a bowl from a drawer and an apple from a refrigerator and placing them on a table; we evaluate seven skills covering picking, placement, refrigerator opening, and drawer opening and closing.
TidyHouse rearranges objects among open receptacles, whereas PrepareGroceries transfers objects between a refrigerator and a counter; both cover picking and placement across nine object categories.
One policy is trained per suite for 100k steps with a batch size of 64, using head- and wrist-camera RGB images, proprioception, and language instructions as observations for every control step.

\Cref{tab:mshab_settable,tab:mshab_tidy} show that MM-ABC achieves the highest mean success rate on SetTable (86.4\%), TidyHouse (70.2\%), and PrepareGroceries (65.6\%).
Improvements on TidyHouse and PrepareGroceries are concentrated in picking.
This pattern is consistent with MM-APT's multilevel perceptual conditioning and coupled manipulation--body prediction, which target object acquisition across varied geometries.

\begin{table*}[t]
    \centering
    \small
    \setlength{\tabcolsep}{4.6pt}
    \caption{
        \textbf{ManiSkill-HAB SetTable}~\citep{chen2026mopa}.
        Skill and mean success rates (\%).
        $^\ast$ denotes depth input; -- indicates an unavailable result.
        AnchorVLA's mean covers six available skills.
        Best: shaded; second best: bold.
    }
    \label{tab:mshab_settable}
    \begin{tabular}{lcccccccc}
        \toprule
        \mmabctablehead & Pick & Pick & Place & Place & Open & Open & Close & \\
        \mmabctablehead Method & Apple & Bowl & Apple & Bowl & Fridge & Drawer & Drawer & Mean \\
        \midrule
        DP3$^\ast$~\citep{ze2024dp3} & 0.0 & 20.0 & 31.0 & 32.0 & 0.0 & 0.0 & 68.0 & 21.6 \\
        ACT~\citep{zhao2023act} & 28.0 & 28.0 & 8.7 & 13.0 & 2.0 & 0.0 & 85.7 & 23.6 \\
        DP~\citep{chi2023diffusionpolicy} & 21.3 & 20.7 & 28.0 & 69.3 & 7.3 & 0.0 & 55.0 & 28.8 \\
        RDT-1B~\citep{liu2024rdt} & 12.0 & 10.7 & 32.0 & 18.7 & 82.7 & 44.0 & \best{100.0} & 42.9 \\
        AC-DiT$^\ast$~\citep{chen2025acdit} & 33.3 & 36.0 & 33.3 & 17.3 & 90.7 & 81.3 & \secondbest{97.3} & 55.6 \\
        $\pi_0$~\citep{black2024pi0} & 26.6 & 26.6 & 48.9 & 56.8 & 90.5 & 75.5 & 88.7 & 59.1 \\
        AnchorVLA~\citep{lim2026anchorvla} & 22.7 & 44.5 & 64.3 & 63.8 & 88.9 & -- & \best{100.0} & 64.0 \\
        MobileWAM~\citep{fan2026mobilewam} & 46.0 & 46.0 & 63.7 & 64.7 & \best{99.3} & \secondbest{91.0} & \best{100.0} & 73.0 \\
        GeoHAT$^\ast$~\citep{zhu2026geohat} & \best{82.3} & 69.3 & 60.0 & 78.0 & 83.7 & 86.0 & 95.3 & 79.2 \\
        InCoM$^\ast$~\citep{liu2026incom} & 59.4 & \secondbest{84.1} & \best{84.1} & \best{82.5} & 87.3 & 88.9 & \best{100.0} & \secondbest{83.8} \\
        \midrule
        \rowcolor{mmabcRow}
        \textbf{MM-ABC (Ours)} & \secondbest{81.3} & \best{86.7} & \secondbest{71.0} & \secondbest{81.3} & \secondbest{96.7} & \best{92.0} & 95.7 & \best{86.4} \\
        \bottomrule
    \end{tabular}
\end{table*}

\begin{table}[t]
    \centering
    \small
    \setlength{\tabcolsep}{5pt}
    \caption{
        \textbf{ManiSkill-HAB TidyHouse and PrepareGroceries}~\citep{chen2026mopa}.
        Pick and Place average success rates (\%) over nine object categories; Mean averages both groups.
        Best: shaded; second best: bold.
    }
    \label{tab:mshab_tidy}
    \begin{tabular}{lcccccc}
        \toprule
        & \multicolumn{3}{c}{TidyHouse} & \multicolumn{3}{c}{PrepareGroceries} \\
        \cmidrule(lr){2-4}\cmidrule(lr){5-7}
        \mmabctablehead Method & Pick & Place & Mean & Pick & Place & Mean \\
        \midrule
        ACT~\citep{zhao2023act} & 2.2 & 31.6 & 16.9 & 2.0 & 27.5 & 14.8 \\
        DP~\citep{chi2023diffusionpolicy} & 0.0 & 30.3 & 15.2 & 0.4 & 17.7 & 9.1 \\
        DP3~\citep{ze2024dp3} & 0.0 & 61.0 & 30.5 & 0.0 & 31.3 & 15.7 \\
        InCoM~\citep{liu2026incom} & 16.7 & \best{78.9} & 47.8 & 15.0 & \secondbest{65.9} & \secondbest{40.5} \\
        GeoHAT~\citep{zhu2026geohat} & \secondbest{30.3} & \secondbest{73.3} & \secondbest{51.8} & \secondbest{19.7} & 60.7 & 40.2 \\
        \midrule
        \rowcolor{mmabcRow}
        \textbf{MM-ABC (Ours)} & \best{71.5} & 68.8 & \best{70.2} & \best{64.4} & \best{66.8} & \best{65.6} \\
        \bottomrule
    \end{tabular}
\end{table}

\textbf{LIBERO and LIBERO-Plus.}
We evaluate LIBERO on four fixed-base manipulation suites---Spatial, Object, Goal, and Long---with ten tasks per suite~\citep{liu2023libero}.
Spatial, Object, and Goal vary object arrangements, object identities, and task objectives, respectively, while Long emphasizes extended sequences of interactions.
We use the standard LIBERO demonstrations for post-training and an evaluation budget of 50 trials per task.
LIBERO-Plus introduces perturbations to camera configuration, robot initial state, language, lighting, background, sensor noise, and scene layout, with 10{,}030 evaluation episodes in total~\citep{fei2025liberoplus}.
For post-training, MM-ABC uses only the original LIBERO demonstrations, without additional training on perturbed demonstrations or environments.

As shown in \cref{tab:libero}, MM-ABC achieves 99.1\% mean success on LIBERO, 0.5 percentage points above the strongest baseline, ABot-M0.
It achieves the highest success rate on Spatial, ties for the highest on Object, and is within 0.4 and 0.1 points of the best results on Goal and Long, respectively.
With the inactive body stream masked (\cref{sec:mmjit}), these results show that the shared perceptual representation and clean-action decoder also support fine-grained fixed-base manipulation.

\Cref{tab:liberoplus} evaluates the LIBERO-trained MM-ABC policy on LIBERO-Plus without further training.
MM-ABC achieves a total success rate of 82.8\%, the highest among the compared methods.
It exceeds Cosmos-Policy (82.2\%) and ABot-M0 (80.5\%), which are also trained only on the original demonstrations, as well as OpenVLA-OFT+ (79.6\%) and GR00T-N1.6+ (79.4\%), which are additionally trained on perturbed demonstrations.
MM-ABC also achieves the highest language (88.9\%) and background (96.1\%) scores, consistent with conditioning actions on shared multilevel visual--language features.
Its lowest category scores occur under camera (73.6\%) and robot initial-state (66.5\%) perturbations.

\begin{table*}[t]
    \centering
    \small
    \setlength{\tabcolsep}{5pt}
    \caption{
        \textbf{LIBERO.}
        Success rate (\%) with 50 trials per task.
        Baselines include suite-specific and shared-policy configurations.
        Best: shaded; second best: bold.
    }
    \label{tab:libero}
    \begin{tabular}{lccccc}
        \toprule
        \mmabctablehead Method & Spatial & Object & Goal & Long & Average \\
        \midrule
        OpenVLA~\citep{kim2024openvla} & 84.7 & 88.4 & 79.2 & 53.7 & 76.5 \\
        WorldVLA~\citep{cen2025worldvla} & 87.6 & 96.2 & 83.4 & 60.0 & 81.8 \\
        $\pi_0$-FAST~\citep{pertsch2025fast} & 96.4 & 96.8 & 88.6 & 60.2 & 85.5 \\
        NORA~\citep{hung2025nora} & 92.2 & 95.4 & 89.4 & 74.6 & 87.9 \\
        $\pi_0$~\citep{black2024pi0} & 96.8 & 98.8 & 95.8 & 85.2 & 94.2 \\
        UniVLA~\citep{bu2025univla} & 96.5 & 96.8 & 95.6 & 92.0 & 95.2 \\
        $\pi_{0.5}$~\citep{pi2025pi05} & 98.8 & 98.2 & 98.0 & 92.4 & 96.9 \\
        OpenVLA-OFT~\citep{kim2025openvlaoft} & 97.6 & 98.4 & 97.9 & 94.5 & 97.1 \\
        GR00T-N1.6~\citep{nvidia2025groot} & \secondbest{99.3} & 99.2 & 98.4 & 92.9 & 97.5 \\
        Fast-WAM~\citep{yuan2026fastwam} & 98.2 & \best{100.0} & 97.0 & 95.2 & 97.6 \\
        StarVLA~\citep{ye2026starvla} & 98.7 & 99.7 & \secondbest{98.6} & 94.2 & 97.8 \\
        X-VLA~\citep{zheng2025xvla} & 98.2 & 98.6 & 97.8 & 97.6 & 98.1 \\
        Cosmos-Policy~\citep{kim2026cosmospolicy} & 98.1 & \best{100.0} & 98.2 & 97.6 & 98.5 \\
        LingBot-VA~\citep{li2026lingbotva} & 98.5 & 99.6 & 97.2 & \best{98.5} & 98.5 \\
        ABot-M0~\citep{yang2026abotm0} & 98.8 & \secondbest{99.8} & \best{99.0} & 96.6 & \secondbest{98.6} \\
        \midrule
        \rowcolor{mmabcRow}
        \textbf{MM-ABC (Ours)} & \best{99.4} & \best{100.0} & \secondbest{98.6} & \secondbest{98.4} & \best{99.1} \\
        \bottomrule
    \end{tabular}
\end{table*}

\begin{table*}[t]
    \centering
    \small
    \setlength{\tabcolsep}{3.8pt}
    \caption{
        \textbf{LIBERO-Plus.}
        Success rate (\%) across seven perturbation categories~\citep{fei2025liberoplus}.
        Total is the success rate over all 10{,}030 evaluation episodes, following the official protocol.
        + denotes additional training on LIBERO-Plus perturbed demonstrations.
        Rankings span both groups.
        Best: shaded; second best: bold.
    }
    \label{tab:liberoplus}
    \begin{tabular}{lcccccccc}
        \toprule
        \mmabctablehead Method & Camera & Robot & Language & Light & Background & Noise & Layout & Total \\
        \midrule
        \multicolumn{9}{l}{\textit{Additional training on perturbed demonstrations}} \\
        $\pi_0$+~\citep{black2024pi0} & 79.6 & 21.1 & 72.5 & 84.7 & 86.2 & 68.3 & 69.4 & 67.4 \\
        GR00T-N1.6+~\citep{nvidia2025groot} & \secondbest{92.6} & 33.5 & 80.1 & 93.6 & 95.4 & \best{93.6} & 75.0 & 79.4 \\
        OpenVLA-OFT+~\citep{fei2025liberoplus} & \best{92.8} & 30.3 & 85.8 & 94.9 & 93.9 & 89.3 & 77.6 & 79.6 \\
        \midrule
        \multicolumn{9}{l}{\textit{Trained on original LIBERO only}} \\
        OpenVLA~\citep{kim2024openvla} & 0.8 & 3.5 & 23.0 & 8.1 & 34.8 & 15.2 & 28.5 & 15.6 \\
        GR00T-N1.6~\citep{nvidia2025groot} & 20.9 & 40.2 & 35.0 & 65.4 & 76.3 & 27.8 & 51.3 & 42.8 \\
        UniVLA~\citep{bu2025univla} & 1.8 & 46.2 & 69.6 & 69.0 & 81.0 & 21.2 & 31.9 & 42.9 \\
        Fast-WAM~\citep{yuan2026fastwam} & 16.4 & 44.5 & 68.9 & 78.2 & 53.7 & 37.7 & 60.7 & 50.0 \\
        $\pi_0$~\citep{black2024pi0} & 13.8 & 6.0 & 58.8 & 85.0 & 81.4 & 79.0 & 68.9 & 53.6 \\
        $\pi_0$-FAST~\citep{pertsch2025fast} & 65.1 & 21.6 & 61.0 & 73.2 & 73.2 & 74.4 & 68.8 & 61.6 \\
        LingBot-VA~\citep{li2026lingbotva} & 40.9 & \secondbest{83.0} & 86.4 & 82.3 & 53.1 & 64.4 & 76.2 & 69.5 \\
        OpenVLA-OFT~\citep{kim2025openvlaoft} & 56.4 & 31.9 & 79.5 & 88.7 & 93.3 & 75.8 & 74.2 & 69.6 \\
        X-VLA~\citep{zheng2025xvla} & 23.4 & \best{89.7} & 75.7 & 88.2 & \secondbest{96.0} & 62.7 & 71.8 & 70.5 \\
        StarVLA~\citep{ye2026starvla} & 52.5 & 49.8 & \secondbest{88.5} & 95.7 & 95.7 & 73.0 & 76.9 & 74.1 \\
        ABot-M0~\citep{yang2026abotm0} & 60.4 & 67.9 & 86.4 & \secondbest{96.2} & 91.6 & 86.4 & \best{82.6} & 80.5 \\
        Cosmos-Policy~\citep{kim2026cosmospolicy} & 75.8 & 63.3 & 81.7 & \best{96.5} & 88.9 & \secondbest{92.7} & \secondbest{82.2} & \secondbest{82.2} \\
        \midrule
        \rowcolor{mmabcRow}
        \textbf{MM-ABC (Ours)} & 73.6 & 66.5 & \best{88.9} & 94.9 & \best{96.1} & 86.4 & 80.5 & \best{82.8} \\
        \bottomrule
    \end{tabular}
\end{table*}

% ------------------------------------------------------------------------------
\subsection{Ablation Studies}
\label{sec:ablations}

\paragraph{Experimental setup.}
We examine the model components on the 16 RoboCasa365 composite-seen tasks.
All variants omit pretraining on our robot-data mixture and use a common training budget of 120k steps with a batch size of 64.
This separates component comparisons from the pretraining in \cref{tab:robocasa365}.

\paragraph{Model variants.}
We compare five configurations.
\textbf{Interleaved} conditions alternating action layers on VLM features, replacing the sparse multilevel conditioning used by the full model.
\textbf{Last layer} removes DeepStack and conditions only on the final VLM representation.
\textbf{$x$-pred} retains clean-action prediction but removes the future supervision loss.
\textbf{$v$-pred} retains the future branch but replaces clean-action prediction with velocity prediction.
The full \textbf{MM-ABC} configuration combines clean-action prediction, future geometric supervision, and DeepStack.
\Cref{tab:ablation} reports per-task success rates and their unweighted mean.

\begin{table*}[t]
    \centering
    \small
    \setlength{\tabcolsep}{6pt}
    \caption{
        \textbf{Component ablations on RoboCasa365 composite-seen tasks.}
        Success rate (\%) without robot-data pretraining.
        Average is the unweighted mean across 16 tasks.
        Rankings are computed within each row, across the five variants.
        Best: shaded; second best: bold.
    }
    \label{tab:ablation}
    \begin{tabular}{lccccc}
        \toprule
        \mmabctablehead Task &
        Interleaved &
        Last layer &
        $x$-pred &
        $v$-pred &
        MM-ABC \\
        \midrule
        DeliverStraw & \secondbest{0.0} & \best{0.5} & \best{0.5} & \best{0.5} & \secondbest{0.0} \\
        GetToastedBread & 0.0 & \best{1.0} & 0.0 & \best{1.0} & \secondbest{0.5} \\
        KettleBoiling & 33.0 & 29.5 & 35.5 & \secondbest{41.5} & \best{49.0} \\
        LoadDishwasher & 18.5 & \secondbest{28.0} & 22.0 & 27.0 & \best{33.0} \\
        PackIdentical & 0.5 & 1.5 & 1.5 & \best{7.0} & \secondbest{6.0} \\
        PreSoakPan & 29.5 & 50.5 & 48.5 & \secondbest{52.0} & \best{67.5} \\
        PrepareCoffee & 13.0 & \secondbest{17.5} & 15.0 & \best{22.0} & \secondbest{17.5} \\
        RinseSinkBasin & 53.5 & 54.5 & \best{62.0} & \secondbest{59.0} & \secondbest{59.0} \\
        ScrubCuttingBoard & 24.5 & \best{47.0} & 19.5 & 20.0 & \secondbest{28.5} \\
        SearingMeat & 10.5 & 3.5 & 8.5 & \secondbest{11.5} & \best{23.0} \\
        SetUpCuttingStation & 21.5 & 14.0 & 18.0 & \secondbest{28.0} & \best{31.0} \\
        StackBowlsCabinet & 57.5 & \secondbest{69.0} & 63.5 & 67.5 & \best{69.5} \\
        SteamInMicrowave & 24.5 & 12.5 & 10.0 & \secondbest{27.0} & \best{29.5} \\
        StirVegetables & \secondbest{19.0} & 5.5 & 12.5 & 15.5 & \best{24.5} \\
        StoreLeftovers & 11.5 & 28.5 & 28.5 & \secondbest{34.5} & \best{48.0} \\
        WashLettuce & 35.5 & \secondbest{46.0} & \best{53.0} & \best{53.0} & 38.0 \\
        \midrule
        \rowcolor{mmabcRow}
        \textbf{Average} & 22.0 & 25.6 & 24.9 & \secondbest{29.2} & \best{32.8} \\
        \bottomrule
    \end{tabular}
\end{table*}

\paragraph{Action prediction and future supervision.}
The full model achieves the highest average success rate, 32.8\%, and the highest success on 9 of the 16 tasks.
Replacing $x$-prediction with $v$-prediction while retaining the future branch reduces the average to 29.2\%, a difference of 3.6 percentage points.
This advantage of clean-action prediction agrees with the controlled analysis in \cref{sec:toy}, which favors the clean endpoint as a prediction target for high-noise denoising and few-step sampling.
Removing future supervision while retaining $x$-prediction reduces the average to 24.9\%, a difference of 7.9 points from the full model.
Because future queries are isolated from action tokens and removed at inference (\cref{sec:future}), this comparison supports the value of auxiliary geometric supervision for the shared perceptual representation.

\paragraph{Perceptual conditioning.}
Final-layer and interleaved conditioning achieve average success rates of 25.6\% and 22.0\%, respectively, compared with 32.8\% for the full model.
These results favor sparse feature injection over the two alternative interfaces, consistent with retaining intermediate visual information alongside a common semantic context (\cref{sec:deepstack}).
At the task level, final-layer conditioning achieves 47.0\% on ScrubCuttingBoard, compared with 28.5\% for the full model; $x$-pred and $v$-pred each achieve 53.0\% on WashLettuce, compared with 38.0\% for the full model.

% ------------------------------------------------------------------------------
\subsection{Real-World Experiments}
\label{sec:real}

\begin{figure}[tp]
    \centering
    \includegraphics[width=\linewidth]{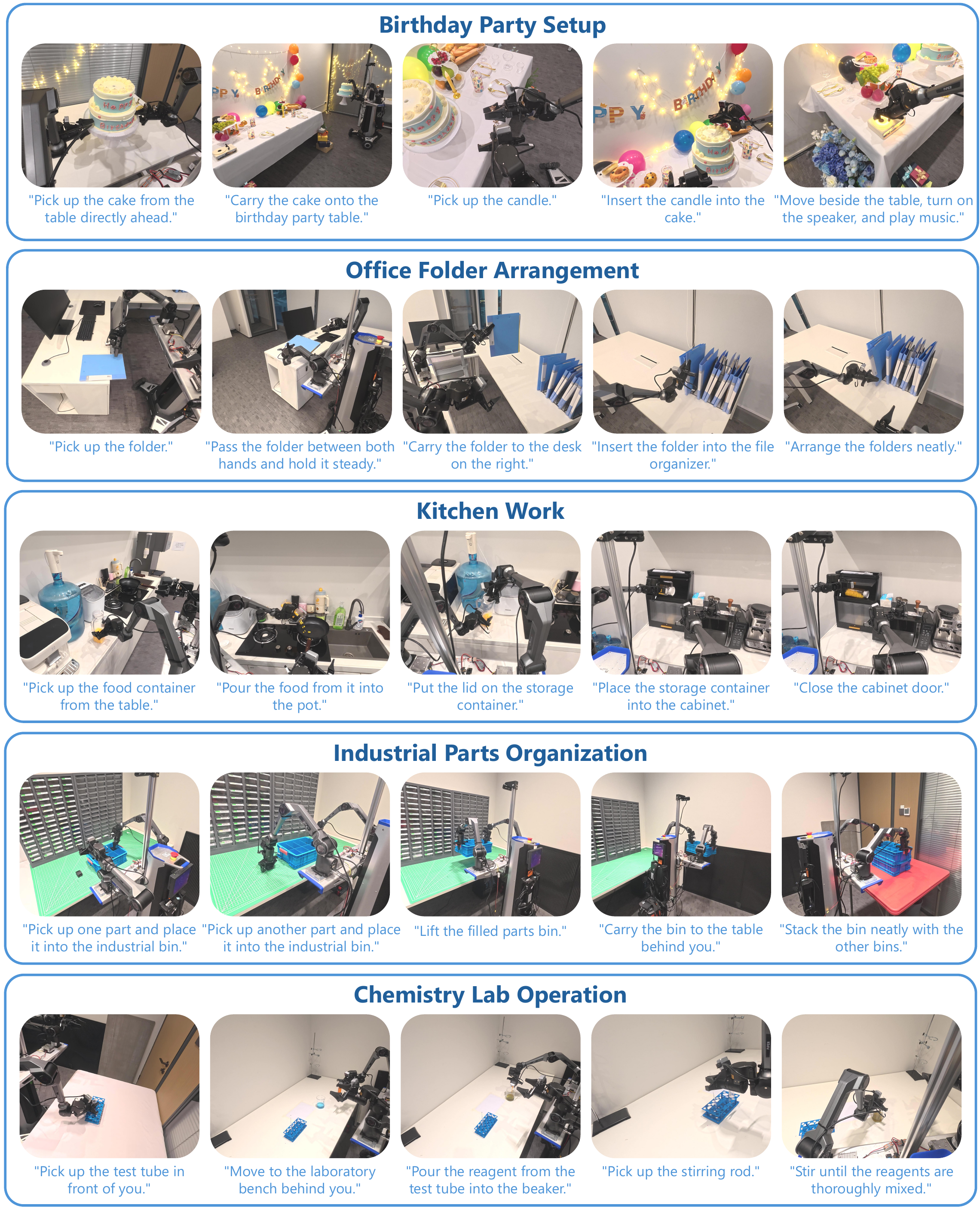}
    \vspace{-0.4em}
    \caption{
        \textbf{Real-world task execution.}
        Representative stages of the five tasks, from top to bottom: Birthday Party Setup,
        Office Folder Arrangement, Kitchen Work, Industrial Parts Organization, and
        Chemistry Lab Operation. Within each task, frames progress from left to right.
    }
    \label{fig:real_tasks}
\end{figure}

\paragraph{Platform.}
We use the mobile manipulation platform described in \cref{sec:selfcollect}, comprising a HexFellow Trigger-A3 omnidirectional base and two AgileX PiPER-X 6-DoF arms.

\paragraph{Tasks.}
The evaluation covers five long-horizon tasks in household, office, workcell, and laboratory scenes.
Each task is specified by a single natural-language instruction and requires base repositioning between object interactions.
\Cref{fig:real_tasks} shows representative stages with step-level instructions.
\begin{itemize}
    \item \textbf{Birthday Party Setup.} The robot picks up a cake from the table directly ahead, carries it to the decorated table on the left, and sets it down. It then picks up a birthday candle, passes it between its two hands, and inserts it into the cake. Finally, it moves beside the table and turns on the speaker to play music.
    \item \textbf{Office Folder Arrangement.} The robot picks up a folder from the office desk in front of it, passes it between both hands, and holds it steady. It then carries the folder to the desk on its right, aligns it with the file organizer, and inserts it.
    \item \textbf{Kitchen Work.} The robot picks up a food container from the table and pours the food into a pot. It then puts the lid on the storage container, places the container into the cabinet, and closes the cabinet door.
    \item \textbf{Industrial Parts Organization.} The robot places scattered industrial parts neatly into a parts bin, carries the filled bin to the table behind it, and stacks it neatly with the other bins.
    \item \textbf{Chemistry Lab Operation.} The robot picks up a test tube, moves to the laboratory bench behind it, pours the reagent into a beaker that already holds another reagent, and stirs until the two are thoroughly mixed.
\end{itemize}

\paragraph{Evaluation protocol.}
We compare MM-ABC with $\pi_{0.5}$~\citep{pi2025pi05} and StarVLA-GR00T~\citep{ye2026starvla}, fine-tuning every method on the same 200 teleoperated demonstrations per task.
Each method is evaluated over 20 trials per task, and a trial counts as successful only when all steps of the task are completed.
We report the percentage of successful trials for each task and the unweighted mean across the five tasks.

\begin{figure}[t]
    \centering
    \includegraphics[width=0.92\linewidth]{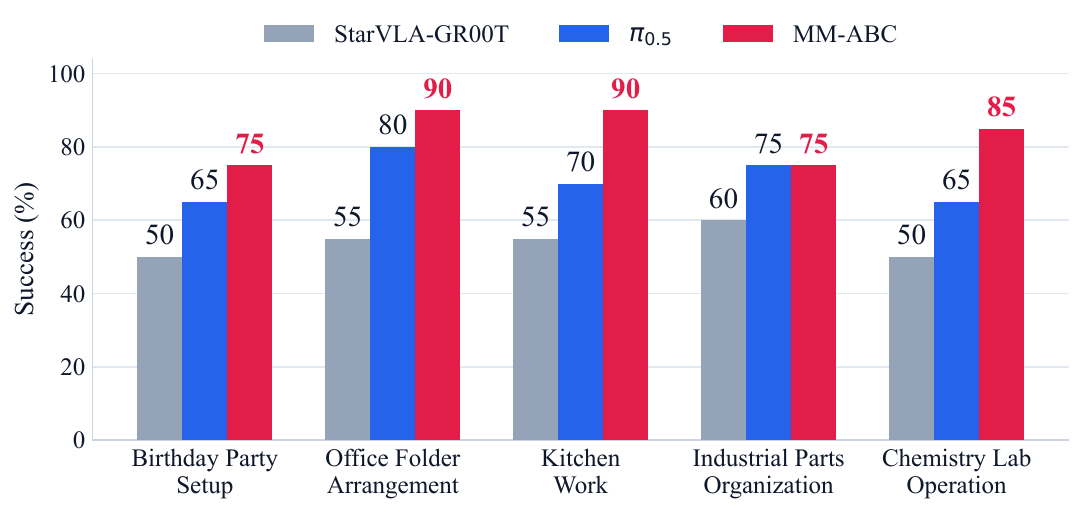}
    \vspace{-0.2em}
    \caption{
        \textbf{Real-world task success.}
        Success rate (\%) over 20 trials per task for each method.
    }
    \label{fig:real_success}
\end{figure}

\paragraph{Results.}
As shown in \cref{fig:real_success}, MM-ABC achieves success rates of 75\%, 90\%, 90\%, 75\%, and 85\% on the five tasks in the order listed above, with a mean of 83\%.
The corresponding rates are 65\%, 80\%, 70\%, 75\%, and 65\% for $\pi_{0.5}$ (mean 71\%), and 50\%, 55\%, 55\%, 60\%, and 50\% for StarVLA-GR00T (mean 54\%).
MM-ABC therefore exceeds the two baselines by 12 and 29 percentage points on average, respectively.
The largest margins over $\pi_{0.5}$, 20 points each, occur on Kitchen Work and Chemistry Lab Operation, which combine base repositioning with pouring, lid placement, cabinet-door closing, and stirring.
On Industrial Parts Organization, MM-ABC and $\pi_{0.5}$ both succeed in 75\% of trials.
% ==============================================================================
% 7. Conclusion
% ==============================================================================
\FloatBarrier
\section{Conclusion}
\label{sec:conclusion}

We presented MM-ABC, a mobile manipulation foundation model built around seeing, coordinating, and imagining arm--base collaboration.
Sparse DeepStack conditioning supplies multilevel VLM features to the action expert, and a training-only future branch aligns the shared perceptual representation with future VGGT-$\Omega$ geometry without adding inference cost.
MM-APT keeps manipulation and body motion in separate streams, couples them through near--far masked joint attention, and predicts clean action chunks.
In a controlled synthetic study, clean-action prediction retains less noise in its endpoint estimates, denoises more accurately at high noise, and allocates the arm--body task more accurately under few-step sampling.

Pretrained on more than 5,000 hours from 12 datasets and 17 embodiments, including the self-collected MM-30 dataset, MM-ABC reaches 44.71\% success on EBench, a task-weighted 61.2\% on RoboCasa365, and the highest mean success on all three ManiSkill-HAB suites.
The same architecture reaches 99.1\% on LIBERO, and its LIBERO-trained policy transfers to LIBERO-Plus with 82.8\% total success.
On five real-world tasks, it averages 83\% success, compared with 71\% for $\pi_{0.5}$ and 54\% for StarVLA-GR00T.
The from-scratch ablation shows that multilevel conditioning, future supervision, and clean-action prediction each contribute: velocity prediction lowers the composite-seen average by 3.6 points, removing future supervision lowers it by 7.9 points, and final-layer or interleaved conditioning trails sparse multilevel injection by 7.2 and 10.8 points.
These results support treating mobile manipulation as manipulation over a reconfigurable workspace, in which perception, anticipation, and arm--base coordination are learned together.

% ==============================================================================
% References (Numbered & Ordered)
% ==============================================================================
\FloatBarrier
\bibliographystyle{unsrtnat}
\bibliography{main}

\begin{thebibliography}{119}
\providecommand{\natexlab}[1]{#1}
\providecommand{\url}[1]{\texttt{#1}}
\expandafter\ifx\csname urlstyle\endcsname\relax
  \providecommand{\doi}[1]{doi: #1}\else
  \providecommand{\doi}{doi: \begingroup \urlstyle{rm}\Url}\fi

\bibitem[Liu et~al.(2026{\natexlab{a}})Liu, Wenbo, Xia, Wang, Li, and Zhao]{liu2026incom}
Jiahao Liu, Cui Wenbo, Zhongpu Xia, Yongliang Wang, Haoran Li, and Dongbin Zhao.
\newblock Incom: Intent-driven perception and structured coordination for mobile manipulation.
\newblock \emph{arXiv preprint arXiv:2602.23024}, 2026{\natexlab{a}}.

\bibitem[Zhu et~al.(2026)Zhu, Wu, Ge, Liu, and Li]{zhu2026geohat}
Xiangyu Zhu, Renjun Wu, Luzhou Ge, Jinyan Liu, and Xuesong Li.
\newblock Geohat: Geometry-adaptive hybrid action transformer for mobile manipulation.
\newblock \emph{arXiv preprint arXiv:2606.13394}, 2026.

\bibitem[Liang et~al.(2026{\natexlab{a}})Liang, Cai, Lai, Zhuang, Lin, Qin, Ye, Liang, and Xu]{liang2026afro}
Qiwei Liang, Boyang Cai, Minghao Lai, Sitong Zhuang, Tao Lin, Yan Qin, Yixuan Ye, Jiaming Liang, and Renjing Xu.
\newblock Bootstrap dynamic-aware 3d visual representation for scalable robot learning.
\newblock In \emph{Proceedings of the IEEE/CVF Conference on Computer Vision and Pattern Recognition}, pages 13419--13429, 2026{\natexlab{a}}.

\bibitem[Tu et~al.(2026)Tu, Shukla, Yoo, Li, Li, Xie, Su, and Tu]{tu2026sgvla}
Ruisen Tu, Arth Shukla, Sohyun Yoo, Xuanlin Li, Junxi Li, Jianwen Xie, Hao Su, and Zhuowen Tu.
\newblock Sg-vla: Learning spatially-grounded vision-language-action models for mobile manipulation.
\newblock \emph{arXiv preprint arXiv:2603.22760}, 2026.

\bibitem[Zheng et~al.(2025)Zheng, Wang, Reed, Bjorck, Fang, Hu, Jang, Kundalia, Lin, Magne, et~al.]{zheng2025flare}
Ruijie Zheng, Jing Wang, Scott Reed, Johan Bjorck, Yu~Fang, Fengyuan Hu, Joel Jang, Kaushil Kundalia, Zongyu Lin, Loic Magne, et~al.
\newblock Flare: Robot learning with implicit world modeling.
\newblock \emph{arXiv preprint arXiv:2505.15659}, 2025.

\bibitem[Chen et~al.(2026{\natexlab{a}})Chen, Yang, Tang, Huo, Lin, Wu, Liu, Chen, Zheng, Yuan, et~al.]{chen2026abotm05}
Ronghan Chen, Yandan Yang, Zuojin Tang, Dongjie Huo, Tong Lin, Haoning Wu, Haoyun Liu, Yuzhi Chen, Lulu Zheng, Botai Yuan, et~al.
\newblock Abot-m0. 5: Unified mobility-and-manipulation world action model.
\newblock \emph{arXiv preprint arXiv:2607.00678}, 2026{\natexlab{a}}.

\bibitem[Li et~al.(2026{\natexlab{a}})Li, Wei, Cao, Wang, Chi, Bai, Sun, Li, Zhang, Jia, et~al.]{li2026wam4d}
Ying Li, Xiaobao Wei, Jiajun Cao, Hao Wang, Xiaowei Chi, Chengyu Bai, Qianpu Sun, Jiajun Li, Xiaojie Zhang, Peidong Jia, et~al.
\newblock Wam4d: Fast 4d world action model via spatial register tokens.
\newblock \emph{arXiv preprint arXiv:2606.14048}, 2026{\natexlab{a}}.

\bibitem[Chen et~al.(2026{\natexlab{b}})Chen, Liu, Qian, Jiang, Liu, Gu, Li, Hou, Wang, Wang, et~al.]{chen2025acdit}
Sixiang Chen, Jiaming Liu, Siyuan Qian, Han Jiang, Zhuoyang Liu, Chenyang Gu, Xiaoqi Li, Chengkai Hou, Pengwei Wang, Zhongyuan Wang, et~al.
\newblock Ac-dit: Adaptive coordination diffusion transformer for mobile manipulation.
\newblock \emph{Advances in Neural Information Processing Systems}, 38:\penalty0 64008--64036, 2026{\natexlab{b}}.

\bibitem[Chen et~al.(2026{\natexlab{c}})Chen, Liang, Zhu, Chen, Xiao, Xie, Zhang, Luo, Xu, and Ding]{chen2026mopa}
Guangyu Chen, Qiwei Liang, Shaolong Zhu, Tianxing Chen, Zikuan Xiao, Yifan Xie, Lingfeng Zhang, Ping Luo, Renjing Xu, and Wenbo Ding.
\newblock Mopa: Coordinated mobile manipulation via subsystem-specific perception alignment.
\newblock \emph{arXiv preprint arXiv:2609.12081}, 2026{\natexlab{c}}.

\bibitem[Khatib(1999)]{khatib1999}
Oussama Khatib.
\newblock Mobile manipulation: The robotic assistant.
\newblock \emph{Robotics and Autonomous Systems}, 26\penalty0 (2-3):\penalty0 175--183, 1999.

\bibitem[Hu et~al.(2023)Hu, Stone, and Mart{\'\i}n-Mart{\'\i}n]{hu2023causal}
Jiaheng Hu, Peter Stone, and Roberto Mart{\'\i}n-Mart{\'\i}n.
\newblock Causal policy gradient for whole-body mobile manipulation.
\newblock \emph{arXiv preprint arXiv:2305.04866}, 2023.

\bibitem[Ehsani et~al.(2024)Ehsani, Gupta, Hendrix, Salvador, Weihs, Zeng, Singh, Kim, Han, Herrasti, et~al.]{ehsani2024spoc}
Kiana Ehsani, Tanmay Gupta, Rose Hendrix, Jordi Salvador, Luca Weihs, Kuo-Hao Zeng, Kunal~Pratap Singh, Yejin Kim, Winson Han, Alvaro Herrasti, et~al.
\newblock Spoc: Imitating shortest paths in simulation enables effective navigation and manipulation in the real world.
\newblock In \emph{2024 IEEE/CVF Conference on Computer Vision and Pattern Recognition (CVPR)}, pages 16238--16250. IEEE, 2024.

\bibitem[Fu et~al.(2024)Fu, Zhao, and Finn]{fu2024mobilealoha}
Zipeng Fu, Tony~Z Zhao, and Chelsea Finn.
\newblock Mobile aloha: Learning bimanual mobile manipulation with low-cost whole-body teleoperation.
\newblock \emph{arXiv preprint arXiv:2401.02117}, 2024.

\bibitem[Jiang et~al.(2025{\natexlab{a}})Jiang, Zhang, Wong, Wang, Ze, Yin, Gokmen, Song, Wu, and Fei-Fei]{jiang2025brs}
Yunfan Jiang, Ruohan Zhang, Josiah Wong, Chen Wang, Yanjie Ze, Hang Yin, Cem Gokmen, Shuran Song, Jiajun Wu, and Li~Fei-Fei.
\newblock Behavior robot suite: Streamlining real-world whole-body manipulation for everyday household activities.
\newblock \emph{arXiv preprint arXiv:2503.05652}, 2025{\natexlab{a}}.

\bibitem[Liang et~al.(2026{\natexlab{b}})Liang, Cai, He, Li, Teng, Duan, Huang, and Zeng]{liang2026legged}
Qiwei Liang, Boyang Cai, Rongyi He, Hui Li, Tao Teng, Haihan Duan, Changxin Huang, and Runhao Zeng.
\newblock Whole-body coordination for dynamic object grasping with legged manipulators.
\newblock In \emph{Proceedings of the AAAI Conference on Artificial Intelligence}, volume~40, pages 18434--18442, 2026{\natexlab{b}}.

\bibitem[Brohan et~al.(2023)Brohan, Brown, Carbajal, Chebotar, Chen, Choromanski, Ding, Driess, Dubey, Finn, et~al.]{brohan2023rt2}
Anthony Brohan, Noah Brown, Justice Carbajal, Yevgen Chebotar, Xi~Chen, Krzysztof Choromanski, Tianli Ding, Danny Driess, Avinava Dubey, Chelsea Finn, et~al.
\newblock Rt-2: Vision-language-action models transfer web knowledge to robotic control.
\newblock \emph{arXiv preprint arXiv:2307.15818}, 2023.

\bibitem[Kim et~al.(2024)Kim, Pertsch, Karamcheti, Xiao, Balakrishna, Nair, Rafailov, Foster, Lam, Sanketi, et~al.]{kim2024openvla}
Moo~Jin Kim, Karl Pertsch, Siddharth Karamcheti, Ted Xiao, Ashwin Balakrishna, Suraj Nair, Rafael Rafailov, Ethan Foster, Grace Lam, Pannag Sanketi, et~al.
\newblock Openvla: An open-source vision-language-action model.
\newblock \emph{arXiv preprint arXiv:2406.09246}, 2024.

\bibitem[Chi et~al.(2025)Chi, Xu, Feng, Cousineau, Du, Burchfiel, Tedrake, and Song]{chi2023diffusionpolicy}
Cheng Chi, Zhenjia Xu, Siyuan Feng, Eric Cousineau, Yilun Du, Benjamin Burchfiel, Russ Tedrake, and Shuran Song.
\newblock Diffusion policy: Visuomotor policy learning via action diffusion.
\newblock \emph{The International Journal of Robotics Research}, 44\penalty0 (10-11):\penalty0 1684--1704, 2025.

\bibitem[Black et~al.(2024)Black, Brown, Driess, Esmail, Equi, Finn, Fusai, Groom, Hausman, Ichter, et~al.]{black2024pi0}
Kevin Black, Noah Brown, Danny Driess, Adnan Esmail, Michael Equi, Chelsea Finn, Niccolo Fusai, Lachy Groom, Karol Hausman, Brian Ichter, et~al.
\newblock $\pi_0$: A vision-language-action flow model for general robot control.
\newblock \emph{arXiv preprint arXiv:2410.24164}, 2024.

\bibitem[Intelligence et~al.(2025)Intelligence, Black, Brown, Darpinian, Dhabalia, Driess, Esmail, Equi, Finn, Fusai, et~al.]{pi2025pi05}
Physical Intelligence, Kevin Black, Noah Brown, James Darpinian, Karan Dhabalia, Danny Driess, Adnan Esmail, Michael Equi, Chelsea Finn, Niccolo Fusai, et~al.
\newblock $\pi_{0.5}$: a vision-language-action model with open-world generalization.
\newblock \emph{arXiv preprint arXiv:2504.16054}, 2025.

\bibitem[Wu et~al.(2025{\natexlab{a}})Wu, Zhou, Xu, Wang, and Yan]{wu2025momanipvla}
Zhenyu Wu, Yuheng Zhou, Xiuwei Xu, Ziwei Wang, and Haibin Yan.
\newblock Momanipvla: Transferring vision-language-action models for general mobile manipulation.
\newblock In \emph{2025 IEEE/CVF Conference on Computer Vision and Pattern Recognition (CVPR)}, pages 1714--1723. IEEE, 2025{\natexlab{a}}.

\bibitem[Team et~al.(2026{\natexlab{a}})Team, Guo, Jin, Li, Li, Li, Liu, Peng, Qin, Su, et~al.]{team2026xiaomi}
Xiaomi~Robotics Team, Jun Guo, Piaopiao Jin, Jason Li, Peiyan Li, Yingyan Li, Futeng Liu, Wanli Peng, Optimus Qin, Yifei Su, et~al.
\newblock Xiaomi-robotics-1: Scaling vision-language-action models with over 100k hours of real-world trajectories.
\newblock \emph{arXiv preprint arXiv:2607.15330}, 2026{\natexlab{a}}.

\bibitem[Wu et~al.(2026)Wu, Wang, Lu, Sun, Liu, Wang, Yan, Wang, Ma, Wang, et~al.]{wu2026foundationapplication}
Wei Wu, Fangjing Wang, Fan Lu, He~Sun, Shi Liu, Yunnan Wang, Yibin Yan, Yong Wang, Shuailei Ma, Xinyang Wang, et~al.
\newblock From foundation to application: Improving vla models in practice.
\newblock \emph{arXiv preprint arXiv:2607.06403}, 2026.

\bibitem[Li et~al.(2025{\natexlab{a}})Li, Deng, Liang, Luo, Zhou, Yao, Zeng, Feng, Liang, Xu, et~al.]{li2025vitra}
Qixiu Li, Yu~Deng, Yaobo Liang, Lin Luo, Lei Zhou, Chengtang Yao, Lingqi Zeng, Zhiyuan Feng, Huizhi Liang, Sicheng Xu, et~al.
\newblock Scalable vision-language-action model pretraining for robotic manipulation with real-life human activity videos.
\newblock \emph{arXiv preprint arXiv:2510.21571}, 2025{\natexlab{a}}.

\bibitem[Wang et~al.(2026{\natexlab{a}})Wang, Li, Guan, Ye, Xie, Liu, Chen, Liang, Zhang, Hu, et~al.]{qwen2026vla}
Qiuyue Wang, Mingsheng Li, Jian Guan, Jinhui Ye, Sicheng Xie, Yitao Liu, Junhao Chen, Zhixuan Liang, Jie Zhang, Xintong Hu, et~al.
\newblock Qwen-vla: Unifying vision-language-action modeling across tasks, environments, and robot embodiments.
\newblock \emph{arXiv preprint arXiv:2605.30280}, 2026{\natexlab{a}}.

\bibitem[Kim et~al.(2026{\natexlab{a}})Kim, Jang, Koo, Jang, Kim, Kim, Yoon, Jang, Choi, Han, et~al.]{kim2026rldx1}
Dongyoung Kim, Huiwon Jang, Myungkyu Koo, Suhyeok Jang, Taeyoung Kim, Beomjun Kim, Byungjun Yoon, Changsung Jang, Daewon Choi, Dongsu Han, et~al.
\newblock Rldx-1 technical report.
\newblock \emph{arXiv preprint arXiv:2605.03269}, 2026{\natexlab{a}}.

\bibitem[Yang et~al.(2026{\natexlab{a}})Yang, Zhang, Chen, Song, Wang, Kang, Wen, Chen, Wang, Xu, et~al.]{yang2026dreamtrajectory}
Zheng Yang, Wenjie Zhang, Xiangyu Chen, Wenxuan Song, Xianpeng Wang, Yihang Kang, Jiawen Wen, Wen Chen, Lujia Wang, Renjing Xu, et~al.
\newblock Dreamtrajectory: Trajectory-guided action generation with world model alignment for mobile manipulation.
\newblock \emph{arXiv preprint arXiv:2608.01381}, 2026{\natexlab{a}}.

\bibitem[Li et~al.(2026{\natexlab{b}})Li, Zhang, Wei, Zhang, Yuan, Zhi, Li, Guo, Gao, Yang, et~al.]{li2026omega0}
Zhe Li, Zhenzhe Zhang, Yangyang Wei, Wenjie Zhang, Xichen Yuan, Peiyuan Zhi, Gen Li, Xinying Guo, Fengjie Gao, Jianfei Yang, et~al.
\newblock $\omega$-0: A latent predictive world action model for concurrent humanoid loco-manipulation.
\newblock \emph{arXiv preprint arXiv:2608.06375}, 2026{\natexlab{b}}.

\bibitem[Esser et~al.(2024)Esser, Kulal, Blattmann, Entezari, M{\"u}ller, Saini, Levi, Lorenz, Sauer, Boesel, et~al.]{esser2024mmdit}
Patrick Esser, Sumith Kulal, Andreas Blattmann, Rahim Entezari, Jonas M{\"u}ller, Harry Saini, Yam Levi, Dominik Lorenz, Axel Sauer, Frederic Boesel, et~al.
\newblock Scaling rectified flow transformers for high-resolution image synthesis.
\newblock In \emph{Forty-first international conference on machine learning}, 2024.

\bibitem[Shridhar et~al.(2023)Shridhar, Manuelli, and Fox]{shridhar2023peract}
Mohit Shridhar, Lucas Manuelli, and Dieter Fox.
\newblock Perceiver-actor: A multi-task transformer for robotic manipulation.
\newblock In \emph{Conference on Robot Learning}, pages 785--799. PMLR, 2023.

\bibitem[Ke et~al.(2024)Ke, Gkanatsios, and Fragkiadaki]{ke20243dda}
Tsung-Wei Ke, Nikolaos Gkanatsios, and Katerina Fragkiadaki.
\newblock 3d diffuser actor: Policy diffusion with 3d scene representations.
\newblock \emph{arXiv preprint arXiv:2402.10885}, 2024.

\bibitem[Li et~al.(2026{\natexlab{c}})Li, Wen, Peng, Peng, and Zhu]{li2025pointvla}
Chengmeng Li, Junjie Wen, Yaxin Peng, Yan Peng, and Yichen Zhu.
\newblock Pointvla: Injecting the 3d world into vision-language-action models.
\newblock \emph{IEEE Robotics and Automation Letters}, 11\penalty0 (3):\penalty0 2506--2513, 2026{\natexlab{c}}.

\bibitem[Sun et~al.(2025)Sun, Xie, Liu, Shi, Wang, and Cao]{sun2025geovla}
Lin Sun, Bin Xie, Yingfei Liu, Hao Shi, Tiancai Wang, and Jiale Cao.
\newblock Geovla: Empowering 3d representations in vision-language-action models.
\newblock \emph{arXiv preprint arXiv:2508.09071}, 2025.

\bibitem[Qu et~al.(2025)Qu, Song, Chen, Yao, Ye, Ding, Wang, Gu, Zhao, Wang, et~al.]{qu2025spatialvla}
Delin Qu, Haoming Song, Qizhi Chen, Yuanqi Yao, Xinyi Ye, Yan Ding, Zhigang Wang, JiaYuan Gu, Bin Zhao, Dong Wang, et~al.
\newblock Spatialvla: Exploring spatial representations for visual-language-action model.
\newblock \emph{arXiv preprint arXiv:2501.15830}, 2025.

\bibitem[Li et~al.(2025{\natexlab{b}})Li, Heng, Liu, Shen, Gu, Liu, Chen, Han, Zhang, Tang, et~al.]{li2025spatial3d}
Xiaoqi Li, Liang Heng, Jiaming Liu, Yan Shen, Chenyang Gu, Zhuoyang Liu, Hao Chen, Nuowei Han, Renrui Zhang, Hao Tang, et~al.
\newblock 3ds-vla: A 3d spatial-aware vision language action model for robust multi-task manipulation.
\newblock In \emph{9th Annual Conference on Robot Learning}, 2025{\natexlab{b}}.

\bibitem[Zhang et~al.(2026{\natexlab{a}})Zhang, Chen, Xu, Huang, Zhou, Yuan, Cai, Huang, Quan, Xu, et~al.]{zhang2025fourdvla}
Jiahui Zhang, Yurui Chen, Yueming Xu, Ze~Huang, Yanpeng Zhou, Yu-Jie Yuan, Xinyue Cai, Guowei Huang, Xingyue Quan, Hang Xu, et~al.
\newblock 4d-vla: Spatiotemporal vision-language-action pretraining with cross-scene calibration.
\newblock \emph{Advances in Neural Information Processing Systems}, 38:\penalty0 33914--33937, 2026{\natexlab{a}}.

\bibitem[Liu et~al.(2026{\natexlab{b}})Liu, Wuwu, Han, Chen, Liu, Fei, Jia, Gu, Guo, Shi, et~al.]{liu2026lift3dvla}
Jiaming Liu, Qingpo Wuwu, Nuowei Han, Hao Chen, Zhuoyang Liu, Fan Fei, Yueru Jia, Chenyang Gu, Yandong Guo, Boxin Shi, et~al.
\newblock Lift3d-vla: Lifting vla models to 3d geometry and dynamics-aware manipulation.
\newblock \emph{arXiv preprint arXiv:2607.06564}, 2026{\natexlab{b}}.

\bibitem[Lin et~al.(2025)Lin, Li, Zhong, Zou, Du, Liu, Gu, and Zhao]{lin2025evo0}
Tao Lin, Gen Li, Yilei Zhong, Yanwen Zou, Yuxin Du, Jiting Liu, Encheng Gu, and Bo~Zhao.
\newblock Evo-0: Vision-language-action model with implicit spatial understanding.
\newblock \emph{arXiv preprint arXiv:2507.00416}, 2025.

\bibitem[Chen et~al.(2026{\natexlab{d}})Chen, Cao, Peng, Zheng, Si, Li, Yan, Zhu, Chen, Fu, et~al.]{chen2026geoalign}
Yizhi Chen, Zhanxiang Cao, Xinyi Peng, Yixiao Zheng, Xiaxi Si, Yiheng Li, Liyun Yan, Keqi Zhu, Xueyun Chen, Shengcheng Fu, et~al.
\newblock Geoalign: Beyond semantics with state-guided spatial alignment in vla models.
\newblock \emph{arXiv preprint arXiv:2606.03240}, 2026{\natexlab{d}}.

\bibitem[Liu et~al.(2026{\natexlab{c}})Liu, Gu, Chen, Zhang, Mao, Wu, Yau, and Wang]{liu2026vistavla}
Mohan Liu, Zhihao Gu, Xuanyu Chen, Haitian Zhang, Kaimin Mao, Yan Wu, Wei-Yun Yau, and Lin Wang.
\newblock Vistavla: Geometry-and semantic-aware 3d gaussian-grounded vla for robotic manipulation.
\newblock \emph{arXiv preprint arXiv:2607.12356}, 2026{\natexlab{c}}.

\bibitem[Bjorck et~al.(2025)Bjorck, Casta{\~n}eda, Cherniadev, Da, Ding, Fan, Fang, Fox, Hu, Huang, et~al.]{nvidia2025groot}
Johan Bjorck, Fernando Casta{\~n}eda, Nikita Cherniadev, Xingye Da, Runyu Ding, Linxi Fan, Yu~Fang, Dieter Fox, Fengyuan Hu, Spencer Huang, et~al.
\newblock Gr00t n1: An open foundation model for generalist humanoid robots.
\newblock \emph{arXiv preprint arXiv:2503.14734}, 2025.

\bibitem[Shukor et~al.(2025)Shukor, Aubakirova, Capuano, Kooijmans, Palma, Zouitine, Aractingi, Pascal, Russi, Marafioti, et~al.]{shukor2025smolvla}
Mustafa Shukor, Dana Aubakirova, Francesco Capuano, Pepijn Kooijmans, Steven Palma, Adil Zouitine, Michel Aractingi, Caroline Pascal, Martino Russi, Andres Marafioti, et~al.
\newblock Smolvla: A vision-language-action model for affordable and efficient robotics.
\newblock \emph{arXiv preprint arXiv:2506.01844}, 2025.

\bibitem[Yang et~al.(2026{\natexlab{b}})Yang, Zeng, Lin, Chang, Qi, Xiao, Liu, Chen, Chen, Huo, et~al.]{yang2026abotm0}
Yandan Yang, Shuang Zeng, Tong Lin, Xinyuan Chang, Dekang Qi, Junjin Xiao, Haoyun Liu, Ronghan Chen, Yuzhi Chen, Dongjie Huo, et~al.
\newblock Abot-m0: Vla foundation model for robotic manipulation with action manifold learning.
\newblock \emph{arXiv preprint arXiv:2602.11236}, 2026{\natexlab{b}}.

\bibitem[Miao et~al.(2026)Miao, Feng, Wu, Lin, He, Li, and Long]{miao2026jepavla}
Shangchen Miao, Ningya Feng, Jialong Wu, Ye~Lin, Xu~He, Dong Li, and Mingsheng Long.
\newblock Jepa-vla: Video predictive embedding is needed for vla models.
\newblock \emph{arXiv preprint arXiv:2602.11832}, 2026.

\bibitem[Meng et~al.(2024)Meng, Yang, Tian, Dai, Wu, Gao, and Jiang]{meng2024deepstack}
Lingchen Meng, Jianwei Yang, Rui Tian, Xiyang Dai, Zuxuan Wu, Jianfeng Gao, and Yu-Gang Jiang.
\newblock Deepstack: Deeply stacking visual tokens is surprisingly simple and effective for lmms.
\newblock \emph{Advances in Neural Information Processing Systems}, 37:\penalty0 23464--23487, 2024.

\bibitem[Bai et~al.(2025)Bai, Cai, Chen, Chen, Chen, Cheng, Deng, Ding, Gao, Ge, et~al.]{bai2025qwen3vl}
Shuai Bai, Yuxuan Cai, Ruizhe Chen, Keqin Chen, Xionghui Chen, Zesen Cheng, Lianghao Deng, Wei Ding, Chang Gao, Chunjiang Ge, et~al.
\newblock Qwen3-vl technical report.
\newblock \emph{arXiv preprint arXiv:2511.21631}, 2025.

\bibitem[Luo et~al.(2026)Luo, Chen, Wu, Sui, Liu, Gu, Liu, Feng, Yu, Gu, et~al.]{luo2026deepvision}
Yulin Luo, Hao Chen, Zhuangzhe Wu, Bowen Sui, Jiaming Liu, Chenyang Gu, Zhuoyang Liu, Qiuxuan Feng, Jiale Yu, Shuo Gu, et~al.
\newblock Look before acting: Enhancing vision foundation representations for vision-language-action models.
\newblock \emph{arXiv preprint arXiv:2603.15618}, 2026.

\bibitem[Yu et~al.(2024)Yu, Kwak, Jang, Jeong, Huang, Shin, and Xie]{yu2025repa}
Sihyun Yu, Sangkyung Kwak, Huiwon Jang, Jongheon Jeong, Jonathan Huang, Jinwoo Shin, and Saining Xie.
\newblock Representation alignment for generation: Training diffusion transformers is easier than you think.
\newblock \emph{arXiv preprint arXiv:2410.06940}, 2024.

\bibitem[Shang et~al.(2024)Shang, Schmeckpeper, May, Minniti, Kelestemur, Watkins, and Herlant]{shang2024theia}
Jinghuan Shang, Karl Schmeckpeper, Brandon~B May, Maria~Vittoria Minniti, Tarik Kelestemur, David Watkins, and Laura Herlant.
\newblock Theia: Distilling diverse vision foundation models for robot learning.
\newblock \emph{arXiv preprint arXiv:2407.20179}, 2024.

\bibitem[Li et~al.(2026{\natexlab{d}})Li, Song, Zhao, Wang, Ding, Wang, Zeng, and Li]{li2026spatialforcing}
Fuhao Li, Wenxuan Song, Han Zhao, Jingbo Wang, Pengxiang Ding, Donglin Wang, Long Zeng, and Haoang Li.
\newblock Spatial forcing: Implicit spatial representation alignment for vision-language-action model.
\newblock In \emph{International Conference on Learning Representations}, volume 2026, pages 132324--132345, 2026{\natexlab{d}}.

\bibitem[Guo et~al.(2025)Guo, Cao, Tao, Xu, Yan, Liang, Laptev, and Chang]{guo2025glad}
Minghao Guo, Meng Cao, Jiachen Tao, Rongtao Xu, Yan Yan, Xiaodan Liang, Ivan Laptev, and Xiaojun Chang.
\newblock Glad: Geometric latent distillation for vision-language-action models.
\newblock \emph{arXiv preprint arXiv:2512.09619}, 2025.

\bibitem[Sun et~al.(2026{\natexlab{a}})Sun, Du, Feng, Luo, Ding, Shen, Wang, He, and Li]{sun2026rocket}
Guoheng Sun, Tingting Du, Kaixi Feng, Chenxiang Luo, Xingguo Ding, Zheyu Shen, Ziyao Wang, Yexiao He, and Ang Li.
\newblock Rocket: Residual-oriented multi-layer alignment for spatially-aware vision-language-action models.
\newblock \emph{arXiv preprint arXiv:2602.17951}, 2026{\natexlab{a}}.

\bibitem[Wang et~al.(2026{\natexlab{b}})Wang, Wei, He, Bai, Fan, Cao, Chen, Li, Rong, Lu, et~al.]{wang2026vega}
Hao Wang, Xiaobao Wei, Jingyang He, Chengyu Bai, Chun-Kai Fan, Jiajun Cao, Jintao Chen, Ying Li, Shanyu Rong, Ming Lu, et~al.
\newblock Vega: Visual encoder grounding alignment for spatially-aware vision-language-action models.
\newblock \emph{arXiv preprint arXiv:2605.10485}, 2026{\natexlab{b}}.

\bibitem[Ding et~al.(2026{\natexlab{a}})Ding, Zhao, Wu, Zhao, Zhao, Zhang, and Cheng]{ding2026mindvla}
Xingyu Ding, Yuzhong Zhao, Yang Wu, Chaoyang Zhao, Chunhai Zhao, Yifan Zhang, and Jian Cheng.
\newblock Mind-vla: Instruction-aware spatial representation alignment for vision-language-action models.
\newblock \emph{arXiv preprint arXiv:2608.04633}, 2026{\natexlab{a}}.

\bibitem[Liu et~al.(2026{\natexlab{d}})Liu, Jie, Sun, Cao, Xu, Liu, and Chen]{liu2026sam3d}
Zonghe Liu, Shanyuan Jie, Xiaoquan Sun, Chen Cao, Zetian Xu, Zongsheng Liu, and Jiayu Chen.
\newblock Sam3d-guided object-centric representation alignment for vision-language-action models.
\newblock \emph{arXiv preprint arXiv:2607.25912}, 2026{\natexlab{d}}.

\bibitem[Li et~al.(2025{\natexlab{c}})Li, Chen, Zhou, Li, Zhang, and Zhao]{li2025qdepth}
Yixuan Li, Yuhui Chen, Mingcai Zhou, Haoran Li, Zhengtao Zhang, and Dongbin Zhao.
\newblock Qdepth-vla: Quantized depth prediction as auxiliary supervision for vision-language-action models.
\newblock \emph{arXiv preprint arXiv:2510.14836}, 2025{\natexlab{c}}.

\bibitem[Song et~al.(2026)Song, Zhou, Zhao, Chen, Ding, Yan, Huang, Tang, Wang, and Li]{song2025reconvla}
Wenxuan Song, Ziyang Zhou, Han Zhao, Jiayi Chen, Pengxiang Ding, Haodong Yan, Yuxin Huang, Feilong Tang, Donglin Wang, and Haoang Li.
\newblock Reconvla: Reconstructive vision-language-action model as effective robot perceiver.
\newblock In \emph{Proceedings of the AAAI Conference on Artificial Intelligence}, volume~40, pages 18549--18557, 2026.

\bibitem[Shi et~al.(2026)Shi, Zhang, Zhu, Li, Zhu, and Yuan]{shi2026think3d}
Jiaxin Shi, Xidong Zhang, Fucai Zhu, Zhe Li, Siyu Zhu, and Weihao Yuan.
\newblock 3dthinkvla: Endowing vision-language-action models with latent 3d priors via 3d-thinking-guided co-training.
\newblock \emph{arXiv preprint arXiv:2606.04436}, 2026.

\bibitem[Lei et~al.(2026)Lei, Jie~Yap, Huang, Xie, Li, Zhang, Liu, and Deng]{tong2026xsvla}
Iok~Tong Lei, Ying Jie~Yap, Wei Huang, Qingchen Xie, Qianzhi Li, Yujie Zhang, Xiaolong Liu, and Zhidong Deng.
\newblock Teaching tiny vla models where to look and how to move.
\newblock \emph{arXiv e-prints}, pages arXiv--2607, 2026.

\bibitem[Li et~al.(2026{\natexlab{e}})Li, Li, Yang, Dong, Rouxel, and Chen]{li2026lss}
Andrew Ting~Yan Li, Zhuo Li, Zhelin Yang, Zhipeng Dong, Quentin Rouxel, and Fei Chen.
\newblock Reasoning without inference cost: Latent semantic scaffolding for robot vla policies.
\newblock \emph{arXiv preprint arXiv:2609.04893}, 2026{\natexlab{e}}.

\bibitem[Liu et~al.(2026{\natexlab{e}})Liu, Jia, Huang, Zhang, and Huang]{liu2026lara}
Mengya Liu, Baoxiong Jia, Jiangyong Huang, Jingze Zhang, and Siyuan Huang.
\newblock Lara: Latent action representation alignment for vision-language-action models.
\newblock \emph{arXiv preprint arXiv:2606.07100}, 2026{\natexlab{e}}.

\bibitem[Ding et~al.(2026{\natexlab{b}})Ding, Zhao, Zhao, Shi, Zhao, and Zhang]{ding2026temporalforcing}
Xingyu Ding, Yuzhong Zhao, Chunhai Zhao, Yinghuan Shi, Chaoyang Zhao, and Yifan Zhang.
\newblock Temporal forcing: 4d representation alignment for vision-language-action models.
\newblock \emph{arXiv preprint arXiv:2608.30643}, 2026{\natexlab{b}}.

\bibitem[Wu et~al.(2024)Wu, Jing, Cheang, Chen, Xu, Li, Liu, Li, and Kong]{wu2024gr1}
Hongtao Wu, Ya~Jing, Chilam Cheang, Guangzeng Chen, Jiafeng Xu, Xinghang Li, Minghuan Liu, Hang Li, and Tao Kong.
\newblock Unleashing large-scale video generative pre-training for visual robot manipulation.
\newblock In \emph{International Conference on Learning Representations}, volume 2024, pages 10641--10662, 2024.

\bibitem[Assran et~al.(2025)Assran, Bardes, Fan, Garrido, Howes, Muckley, Rizvi, Roberts, Sinha, Zholus, et~al.]{assran2025vjepa2}
Mido Assran, Adrien Bardes, David Fan, Quentin Garrido, Russell Howes, Matthew Muckley, Ammar Rizvi, Claire Roberts, Koustuv Sinha, Artem Zholus, et~al.
\newblock V-jepa 2: Self-supervised video models enable understanding, prediction and planning.
\newblock \emph{arXiv preprint arXiv:2506.09985}, 2025.

\bibitem[Sun et~al.(2026{\natexlab{b}})Sun, Zhang, Qi, Ren, Liu, Zhu, Sun, Jin, and Chen]{sun2026vlajepa}
Jingwen Sun, Wenyao Zhang, Zekun Qi, Shaojie Ren, Zezhi Liu, Hanxin Zhu, Guangzhong Sun, Xin Jin, and Zhibo Chen.
\newblock Vla-jepa: Enhancing vision-language-action model with latent world model.
\newblock In \emph{European Conference on Computer Vision}, pages 478--497. Springer, 2026{\natexlab{b}}.

\bibitem[Zhang et~al.(2026{\natexlab{b}})Zhang, Liu, Qi, Wang, Yu, Zhang, Dong, He, Wang, Zhang, et~al.]{zhang2025dreamvla}
Wenyao Zhang, Hongsi Liu, Zekun Qi, Yunnan Wang, Xinqiang Yu, Jiazhao Zhang, Runpei Dong, Jiawei He, He~Wang, Zhizheng Zhang, et~al.
\newblock Dreamvla: a vision-language-action model dreamed with comprehensive world knowledge.
\newblock \emph{Advances in Neural Information Processing Systems}, 38:\penalty0 24195--24228, 2026{\natexlab{b}}.

\bibitem[Han et~al.(2026)Han, Jeon, Jung, Zurbr{\"u}gg, An, Portela, Hutter, Pollefeys, Kim, and Hong]{han2026gam}
Jisang Han, Seonghu Jeon, Jaewoo Jung, Ren{\'e} Zurbr{\"u}gg, Honggyu An, Tifanny Portela, Marco Hutter, Marc Pollefeys, Seungryong Kim, and Sunghwan Hong.
\newblock Geometric action model for robot policy learning.
\newblock \emph{arXiv preprint arXiv:2606.17046}, 2026.

\bibitem[Yang et~al.(2026{\natexlab{c}})Yang, Song, Wang, Sheng, Fang, Zhou, He, Yan, Chen, Sun, et~al.]{yang20264dwam}
Lishan Yang, Wenxuan Song, Xi~Wang, Pingyue Sheng, Zheng Fang, Ziyang Zhou, Junjie He, Haodong Yan, Jiayi Chen, Nan Sun, et~al.
\newblock 4d-wam: Infusing spatiotemporal awareness into world action models through trajectory fields.
\newblock \emph{arXiv preprint arXiv:2608.08023}, 2026{\natexlab{c}}.

\bibitem[Soleymanzadeh et~al.(2026)Soleymanzadeh, Zhang, Zhang, Zhang, Liang, She, and Zheng]{soleymanzadeh2026phrvla}
Davood Soleymanzadeh, Kaidi Zhang, Zhiyuan Zhang, Bihao Zhang, Xiao Liang, Yu~She, and Minghui Zheng.
\newblock Phr-vla: Planning horizon reasoning for vision-language-action models.
\newblock \emph{arXiv preprint arXiv:2608.27609}, 2026.

\bibitem[Zhang et~al.(2026{\natexlab{c}})Zhang, Zhu, Su, Ma, Huang, Xu, and Wang]{zhang2026mecowam}
Jianjun Zhang, Jian Zhu, Taiyi Su, Chong Ma, Zitai Huang, Yi~Xu, and Hanli Wang.
\newblock Learning 4d geometric priors for inference-efficient world action models.
\newblock \emph{arXiv preprint arXiv:2607.05468}, 2026{\natexlab{c}}.

\bibitem[Wang et~al.(2026{\natexlab{c}})Wang, Chen, Zhang, Karaev, Sch{\"o}nberger, Labatut, Bojanowski, Novotny, Vedaldi, and Rupprecht]{wang2026vggtomega}
Jianyuan Wang, Minghao Chen, Shangzhan Zhang, Nikita Karaev, Johannes Sch{\"o}nberger, Patrick Labatut, Piotr Bojanowski, David Novotny, Andrea Vedaldi, and Christian Rupprecht.
\newblock Vggt-$\omega$.
\newblock \emph{arXiv preprint arXiv:2605.15195}, 2026{\natexlab{c}}.

\bibitem[Salimans and Ho(2022)]{salimans2022distillation}
Tim Salimans and Jonathan Ho.
\newblock Progressive distillation for fast sampling of diffusion models.
\newblock \emph{arXiv preprint arXiv:2202.00512}, 2022.

\bibitem[Karras et~al.(2022)Karras, Aittala, Aila, and Laine]{karras2022edm}
Tero Karras, Miika Aittala, Timo Aila, and Samuli Laine.
\newblock Elucidating the design space of diffusion-based generative models.
\newblock \emph{Advances in neural information processing systems}, 35:\penalty0 26565--26577, 2022.

\bibitem[Lipman et~al.(2022)Lipman, Chen, Ben-Hamu, Nickel, and Le]{lipman2023flowmatching}
Yaron Lipman, Ricky~TQ Chen, Heli Ben-Hamu, Maximilian Nickel, and Matt Le.
\newblock Flow matching for generative modeling.
\newblock \emph{arXiv preprint arXiv:2210.02747}, 2022.

\bibitem[Peebles and Xie(2023)]{peebles2023dit}
William Peebles and Saining Xie.
\newblock Scalable diffusion models with transformers.
\newblock In \emph{2023 IEEE/CVF International Conference on Computer Vision (ICCV)}, pages 4172--4182. IEEE, 2023.

\bibitem[Ze et~al.(2024)Ze, Zhang, Zhang, Hu, Wang, and Xu]{ze2024dp3}
Yanjie Ze, Gu~Zhang, Kangning Zhang, Chenyuan Hu, Muhan Wang, and Huazhe Xu.
\newblock 3d diffusion policy: Generalizable visuomotor policy learning via simple 3d representations.
\newblock \emph{arXiv preprint arXiv:2403.03954}, 2024.

\bibitem[Liu et~al.(2025)Liu, Wu, Li, Tan, Chen, Wang, Xu, Su, and Zhu]{liu2024rdt}
Songming Liu, Lingxuan Wu, Bangguo Li, Hengkai Tan, Huayu Chen, Zhengyi Wang, Ke~Xu, Hang Su, and Jun Zhu.
\newblock Rdt-1b: a diffusion foundation model for bimanual manipulation.
\newblock In \emph{International Conference on Learning Representations}, volume 2025, pages 29982--30009, 2025.

\bibitem[Gao et~al.(2024)Gao, Lu, Chen, Dai, Wang, Shang, Ding, and Tang]{gao2024manicm}
Zifeng Gao, Guanxing Lu, Tianxing Chen, Wenxun Dai, Ziwei Wang, Chao Shang, Wenbo Ding, and Yansong Tang.
\newblock Manicm: Real-time 3d diffusion policy via consistency model for robotic manipulation.
\newblock \emph{arXiv preprint arXiv:2406.01586}, 2024.

\bibitem[Zhuo et~al.(2026)Zhuo, Chen, Xue, Tang, Lv, Lu, and Wen]{zhuo2026fardp}
Lifeng Zhuo, Wendi Chen, Han Xue, Shirun Tang, Jun Lv, Cewu Lu, and Chuan Wen.
\newblock Fa-rdp: A frequency-adaptive reactive diffusion policy for contact-rich manipulation.
\newblock \emph{arXiv preprint arXiv:2607.28596}, 2026.

\bibitem[Koirala and Campbell(2026)]{koirala2026vgfm}
Prajwal Koirala and Mark Campbell.
\newblock Vgfm: Expressive robot policies via dense value guidance in flow matching.
\newblock \emph{arXiv preprint arXiv:2609.14261}, 2026.

\bibitem[Li and He(2026)]{li2025jit}
Tianhong Li and Kaiming He.
\newblock Back to basics: Let denoising generative models denoise.
\newblock In \emph{Proceedings of the IEEE/CVF Conference on Computer Vision and Pattern Recognition}, pages 36115--36125, 2026.

\bibitem[Lu et~al.(2026)Lu, Lu, Sun, Zhao, Jiang, Wang, Li, Geng, and He]{lu2026pixelmeanflow}
Yiyang Lu, Susie Lu, Qiao Sun, Hanhong Zhao, Zhicheng Jiang, Xianbang Wang, Tianhong Li, Zhengyang Geng, and Kaiming He.
\newblock One-step latent-free image generation with pixel mean flows.
\newblock \emph{arXiv preprint arXiv:2601.22158}, 2026.

\bibitem[Wang et~al.(2026{\natexlab{d}})Wang, Zhao, Lu, Zhou, Ma, and He]{wang2026minit2i}
Xianbang Wang, Hanhong Zhao, Yiyang Lu, Kangyang Zhou, Linrui Ma, and Kaiming He.
\newblock Minit2i: A minimalist baseline for text-to-image generation, 2026{\natexlab{d}}.
\newblock URL \url{https://peppaking8.github.io/#/post/minit2i}.

\bibitem[Pan et~al.(2026)Pan, Anantharaman, Huang, Jin, Pfrommer, Yuan, Permenter, Qu, Boffi, Shi, et~al.]{pan2025noising}
Chaoyi Pan, Giridharan Anantharaman, Nai-Chieh Huang, Claire Jin, Daniel Pfrommer, Chenyang Yuan, Frank Permenter, Guannan Qu, Nicholas Boffi, Guanya Shi, et~al.
\newblock Much ado about noising: Dispelling the myths of generative robotic control.
\newblock In \emph{International Conference on Learning Representations}, volume 2026, pages 90575--90614, 2026.

\bibitem[Merel et~al.(2019)Merel, Botvinick, and Wayne]{merel2019hierarchical}
Josh Merel, Matthew Botvinick, and Greg Wayne.
\newblock Hierarchical motor control in mammals and machines.
\newblock \emph{Nature communications}, 10\penalty0 (1):\penalty0 5489, 2019.

\bibitem[Allshire et~al.(2026)Allshire, Singh, Singh, Rashid, Choi, McAllister, Yu, Chen, Huang, Abbeel, et~al.]{allshire2026abc}
Arthur Allshire, Himanshu~Gaurav Singh, Ritvik Singh, Adam Rashid, Hongsuk Choi, David McAllister, Justin Yu, Yiyuan Chen, Huang Huang, Pieter Abbeel, et~al.
\newblock Scalable behavior cloning with open data, training, and evaluation.
\newblock \emph{arXiv preprint arXiv:2606.27375}, 2026.

\bibitem[Wu et~al.(2025{\natexlab{b}})Wu, Liu, Xie, Wang, Li, Yang, Li, Zhu, Wu, Liu, et~al.]{wu2025robocoin}
Shihan Wu, Xuecheng Liu, Shaoxuan Xie, Pengwei Wang, Xinghang Li, Bowen Yang, Zhe Li, Kai Zhu, Hongyu Wu, Yiheng Liu, et~al.
\newblock Robocoin: An open-sourced bimanual robotic data collection for integrated manipulation.
\newblock \emph{arXiv preprint arXiv:2511.17441}, 2025{\natexlab{b}}.

\bibitem[Li et~al.(2023)Li, Zhang, Wong, Gokmen, Srivastava, Mart{\'\i}n-Mart{\'\i}n, Wang, Levine, Lingelbach, Sun, et~al.]{li2024behavior1k}
Chengshu Li, Ruohan Zhang, Josiah Wong, Cem Gokmen, Sanjana Srivastava, Roberto Mart{\'\i}n-Mart{\'\i}n, Chen Wang, Gabrael Levine, Michael Lingelbach, Jiankai Sun, et~al.
\newblock Behavior-1k: A benchmark for embodied ai with 1,000 everyday activities and realistic simulation.
\newblock In \emph{Conference on Robot Learning}, pages 80--93. PMLR, 2023.

\bibitem[Zhang et~al.(2026{\natexlab{d}})Zhang, Xiang, Lin, Huang, Wang, Zhong, Dong, Wu, Rao, Zhang, et~al.]{zhang2026hyvla}
He~Zhang, Lingzhu Xiang, Haitao Lin, Zeyu Huang, Minghui Wang, Dingyan Zhong, Yubo Dong, Yihao Wu, Yongming Rao, Dongsheng Zhang, et~al.
\newblock Hy-embodied-0.5-vla: From vision-language-action models to a real-world robot learning stack.
\newblock \emph{arXiv preprint arXiv:2606.14409}, 2026{\natexlab{d}}.

\bibitem[Bu et~al.(2025{\natexlab{a}})Bu, Cai, Chen, Cui, Ding, Feng, Gao, He, Hu, Huang, et~al.]{bu2025agibotworld}
Qingwen Bu, Jisong Cai, Li~Chen, Xiuqi Cui, Yan Ding, Siyuan Feng, Shenyuan Gao, Xindong He, Xuan Hu, Xu~Huang, et~al.
\newblock Agibot world colosseo: A large-scale manipulation platform for scalable and intelligent embodied systems.
\newblock \emph{arXiv preprint arXiv:2503.06669}, 2025{\natexlab{a}}.

\bibitem[Tian et~al.(2026)Tian, Yang, Xie, Cai, Shi, Gao, Liu, Jiang, Qiu, Yuan, et~al.]{tian2026interndata}
Yang Tian, Yuyin Yang, Yiman Xie, Zetao Cai, Xu~Shi, Ning Gao, Hangxu Liu, Xuekun Jiang, Zherui Qiu, Feng Yuan, et~al.
\newblock Interndata-a1: Pioneering high-fidelity synthetic data for pre-training generalist policy.
\newblock In \emph{Proceedings of the IEEE/CVF Conference on Computer Vision and Pattern Recognition}, pages 976--985, 2026.

\bibitem[RealSource(2025)]{realsource2025world}
RealSource.
\newblock Realsource world: A large-scale real-world dual-arm manipulation dataset.
\newblock \url{https://huggingface.co/datasets/RealSourceData/RealSource-World}, 2025.

\bibitem[Jiang et~al.(2025{\natexlab{b}})Jiang, Yuan, Liu, Lu, Cui, Liu, Cheng, Gao, Xu, and Zhao]{jiang2025galaxea}
Tao Jiang, Tianyuan Yuan, Yicheng Liu, Chenhao Lu, Jianning Cui, Xiao Liu, Shuiqi Cheng, Jiyang Gao, Huazhe Xu, and Hang Zhao.
\newblock Galaxea open-world dataset and g0 dual-system vla model.
\newblock \emph{arXiv preprint arXiv:2509.00576}, 2025{\natexlab{b}}.

\bibitem[Khazatsky et~al.(2024)Khazatsky, Pertsch, Nair, Balakrishna, Dasari, Karamcheti, Nasiriany, Srirama, Chen, Ellis, et~al.]{khazatsky2024droid}
Alexander Khazatsky, Karl Pertsch, Suraj Nair, Ashwin Balakrishna, Sudeep Dasari, Siddharth Karamcheti, Soroush Nasiriany, Mohan~Kumar Srirama, Lawrence~Yunliang Chen, Kirsty Ellis, et~al.
\newblock Droid: A large-scale in-the-wild robot manipulation dataset.
\newblock \emph{arXiv preprint arXiv:2403.12945}, 2024.

\bibitem[Team(2026)]{agibotworld2026}
AgiBot~World Team.
\newblock Agibot world 2026.
\newblock \url{https://huggingface.co/datasets/agibot-world/AgiBotWorld2026}, 2026.

\bibitem[BitRobot et~al.(2026)BitRobot, Unitree, and Face]{hiw500_2026}
BitRobot, Unitree, and Hugging Face.
\newblock Hiw-500: Humanoids in-the-wild dataset for robot learning.
\newblock \url{https://bitrobot-foundation.github.io/humanoids-in-the-wild-500-hours/}, 2026.

\bibitem[Cai et~al.(2026)Cai, Liang, Li, Weng, Zhang, Lin, Chen, Zhang, Mao, Xu, et~al.]{cai2026multiview}
Boyang Cai, Qiwei Liang, Jiawei Li, Shihang Weng, Zhaoxin Zhang, Tao Lin, Xiangyu Chen, Wenjie Zhang, Jiaqi Mao, Weisheng Xu, et~al.
\newblock Beyond viewpoint generalization: What multi-view demonstrations offer and how to synthesize them for robot manipulation?
\newblock \emph{arXiv preprint arXiv:2603.26757}, 2026.

\bibitem[Community et~al.(2026)Community, Chen, Chen, Nian, Cai, Chen, Lin, Liang, Xiang, Su, et~al.]{community2026xpolicylab}
XPolicyLab Community, Tianxing Chen, Yue Chen, Tian Nian, Zijian Cai, Guangyu Chen, Wenwei Lin, Qiwei Liang, Peicheng Xiang, Kailun Su, et~al.
\newblock Xpolicylab: A unified standard and open ecosystem for robot policy evaluation and deployment.
\newblock \emph{arXiv preprint arXiv:2608.09892}, 2026.

\bibitem[Gao et~al.(2026)Gao, Zheng, Gao, Ma, Wang, Wang, Chen, Chen, Zhang, Jia, et~al.]{gao2026ebench}
Ning Gao, Jinliang Zheng, Xing Gao, Haoxiang Ma, Hanqing Wang, Yukai Wang, Jiantong Chen, Zanxin Chen, Shujie Zhang, Mingda Jia, et~al.
\newblock Ebench: Elemental diagnosis of generalist mobile manipulation policies.
\newblock \emph{arXiv preprint arXiv:2606.18239}, 2026.

\bibitem[Ma et~al.(2026)Ma, Cai, Xu, Li, Yang, Tian, Cao, Zhu, Qiu, Yang, et~al.]{internvla2026a15}
Haoxiang Ma, Junhao Cai, Xiaoxu Xu, Hao Li, Yuyin Yang, Yang Tian, Jiafei Cao, Hongrui Zhu, Zherui Qiu, Yuqiang Yang, et~al.
\newblock Internvla-a1. 5: Unifying understanding, latent foresight, and action for compositional generalization.
\newblock \emph{arXiv preprint arXiv:2607.04988}, 2026.

\bibitem[Team et~al.(2026{\natexlab{b}})Team, Ye, Sun, Jin, Cheng, Shi, Shang, Zhang, Huang, Wang, et~al.]{gigabrain2026}
GigaBrain Team, Angen Ye, Axiang Sun, Can Jin, Chenxi Cheng, Chong Shi, Dengke Shang, Dingqian Zhang, Guan Huang, Guangqiang Wang, et~al.
\newblock Gigabrain-0.7: Scaling embodied foundation models to emergent capabilities with a three-system architecture.
\newblock \emph{arXiv preprint arXiv:2608.15875}, 2026{\natexlab{b}}.

\bibitem[Agarwal et~al.(2026)Agarwal, Ali, Allen, Antolini, Aubame, Azzolini, Bai, Bala, Balaji, Bapst, et~al.]{nvidia2026cosmos3}
Niket Agarwal, Arslan Ali, Jon Allen, Martin Antolini, Adeline Aubame, Alisson Azzolini, Junjie Bai, Maciej Bala, Yogesh Balaji, Josh Bapst, et~al.
\newblock Cosmos 3: Omnimodal world models for physical ai.
\newblock \emph{arXiv preprint arXiv:2606.02800}, 2026.

\bibitem[Yuan et~al.(2026)Yuan, Dong, Liu, and Zhao]{yuan2026fastwam}
Tianyuan Yuan, Zibin Dong, Yicheng Liu, and Hang Zhao.
\newblock Fast-wam: Do world action models need test-time future imagination?
\newblock \emph{arXiv preprint arXiv:2603.16666}, 2026.

\bibitem[Zheng et~al.(2026)Zheng, Li, Wang, Liu, Kang, Feng, Zheng, Zou, Chen, Zeng, et~al.]{zheng2025xvla}
Jinliang Zheng, Jianxiong Li, Zhihao Wang, Dongxiu Liu, Xirui Kang, Yuchun Feng, Yinan Zheng, Jiayin Zou, Yilun Chen, Jia Zeng, et~al.
\newblock X-vla: Soft-prompted transformer as scalable cross-embodiment vision-language-action model.
\newblock In \emph{International Conference on Learning Representations}, volume 2026, pages 60580--60606, 2026.

\bibitem[Nasiriany et~al.(2026)Nasiriany, Nasiriany, Maddukuri, and Zhu]{nasiriany2026robocasa365}
Soroush Nasiriany, Sep Nasiriany, Abhiram Maddukuri, and Yuke Zhu.
\newblock Robocasa365: A large-scale simulation framework for training and benchmarking generalist robots.
\newblock In \emph{International Conference on Learning Representations}, volume 2026, pages 98643--98667, 2026.

\bibitem[Li et~al.(2026{\natexlab{f}})Li, Zhang, Luo, Yang, Wang, Han, Yu, Gao, Xue, Zhu, et~al.]{li2026lingbotva}
Lin Li, Qihang Zhang, Yiming Luo, Shuai Yang, Ruilin Wang, Fei Han, Mingrui Yu, Zelin Gao, Nan Xue, Xing Zhu, et~al.
\newblock Causal world modeling for robot control.
\newblock \emph{arXiv preprint arXiv:2601.21998}, 2026{\natexlab{f}}.

\bibitem[Shukla et~al.(2025)Shukla, Tao, and Su]{shukla2025maniskillhab}
Arth Shukla, Stone Tao, and Hao Su.
\newblock Maniskill-hab: A benchmark for low-level manipulation in home rearrangement tasks.
\newblock In \emph{International Conference on Learning Representations}, volume 2025, pages 15288--15317, 2025.

\bibitem[Zhao et~al.(2023)Zhao, Kumar, Levine, and Finn]{zhao2023act}
Tony~Z Zhao, Vikash Kumar, Sergey Levine, and Chelsea Finn.
\newblock Learning fine-grained bimanual manipulation with low-cost hardware.
\newblock \emph{arXiv preprint arXiv:2304.13705}, 2023.

\bibitem[Lim et~al.(2026)Lim, Zhang, Bohm, Tidd, Huang, and Luo]{lim2026anchorvla}
Jia~Syuen Lim, Zhizhen Zhang, Peter Bohm, Brendan Tidd, Zi~Huang, and Yadan Luo.
\newblock Anchorvla: Anchored diffusion for efficient end-to-end mobile manipulation.
\newblock \emph{arXiv preprint arXiv:2604.01567}, 2026.

\bibitem[Fan et~al.(2026)Fan, He, Song, Wang, Lyu, Zhao, Li, You, Yang, Xu, et~al.]{fan2026mobilewam}
Zehua Fan, Junjie He, Wenxuan Song, Xi~Wang, Wenqi Lyu, Linge Zhao, Fuhao Li, Zihan You, Yifei Yang, Kaiming Xu, et~al.
\newblock Mobilewam: Bridging world action models to mobile manipulation with chain-of-foresight.
\newblock \emph{arXiv preprint arXiv:2608.04657}, 2026.

\bibitem[Liu et~al.(2023)Liu, Zhu, Gao, Feng, Liu, Zhu, and Stone]{liu2023libero}
Bo~Liu, Yifeng Zhu, Chongkai Gao, Yihao Feng, Qiang Liu, Yuke Zhu, and Peter Stone.
\newblock Libero: Benchmarking knowledge transfer for lifelong robot learning.
\newblock \emph{Advances in Neural Information Processing Systems}, 36:\penalty0 44776--44791, 2023.

\bibitem[Fei et~al.(2025)Fei, Wang, Shi, Dai, Cai, Qian, Ji, He, Zhang, Fei, et~al.]{fei2025liberoplus}
Senyu Fei, Siyin Wang, Junhao Shi, Zihao Dai, Jikun Cai, Pengfang Qian, Li~Ji, Xinzhe He, Shiduo Zhang, Zhaoye Fei, et~al.
\newblock Libero-plus: In-depth robustness analysis of vision-language-action models.
\newblock \emph{arXiv preprint arXiv:2510.13626}, 2025.

\bibitem[Cen et~al.(2025)Cen, Yu, Yuan, Jiang, Huang, Guo, Li, Song, Luo, Wang, et~al.]{cen2025worldvla}
Jun Cen, Chaohui Yu, Hangjie Yuan, Yuming Jiang, Siteng Huang, Jiayan Guo, Xin Li, Yibing Song, Hao Luo, Fan Wang, et~al.
\newblock Worldvla: Towards autoregressive action world model.
\newblock \emph{arXiv preprint arXiv:2506.21539}, 2025.

\bibitem[Pertsch et~al.(2025)Pertsch, Stachowicz, Ichter, Driess, Nair, Vuong, Mees, Finn, and Levine]{pertsch2025fast}
Karl Pertsch, Kyle Stachowicz, Brian Ichter, Danny Driess, Suraj Nair, Quan Vuong, Oier Mees, Chelsea Finn, and Sergey Levine.
\newblock Fast: Efficient action tokenization for vision-language-action models.
\newblock \emph{arXiv preprint arXiv:2501.09747}, 2025.

\bibitem[Hung et~al.(2025)Hung, Sun, Hong, Zadeh, Li, Tan, Majumder, Poria, et~al.]{hung2025nora}
Chia-Yu Hung, Qi~Sun, Pengfei Hong, Amir Zadeh, Chuan Li, U~Tan, Navonil Majumder, Soujanya Poria, et~al.
\newblock Nora: A small open-sourced generalist vision language action model for embodied tasks.
\newblock \emph{arXiv preprint arXiv:2504.19854}, 2025.

\bibitem[Bu et~al.(2025{\natexlab{b}})Bu, Yang, Cai, Gao, Ren, Yao, Luo, and Li]{bu2025univla}
Qingwen Bu, Yanting Yang, Jisong Cai, Shenyuan Gao, Guanghui Ren, Maoqing Yao, Ping Luo, and Hongyang Li.
\newblock Univla: Learning to act anywhere with task-centric latent actions.
\newblock \emph{arXiv preprint arXiv:2505.06111}, 2025{\natexlab{b}}.

\bibitem[Kim et~al.(2025)Kim, Finn, and Liang]{kim2025openvlaoft}
Moo~Jin Kim, Chelsea Finn, and Percy Liang.
\newblock Fine-tuning vision-language-action models: Optimizing speed and success.
\newblock \emph{arXiv preprint arXiv:2502.19645}, 2025.

\bibitem[Ye et~al.(2026)Ye, Gao, Yang, Zheng, Wang, Chen, Chen, Chen, Liu, and Jia]{ye2026starvla}
Jinhui Ye, Ning Gao, Senqiao Yang, Jinliang Zheng, Zixuan Wang, Yuxin Chen, Pengguang Chen, Yilun Chen, Shu Liu, and Jiaya Jia.
\newblock Starvla-$\alpha$: Reducing complexity in vision-language-action systems.
\newblock \emph{arXiv preprint arXiv:2604.11757}, 2026.

\bibitem[Kim et~al.(2026{\natexlab{b}})Kim, Gao, Lin, Lin, Ge, Lam, Liang, Song, Liu, Finn, et~al.]{kim2026cosmospolicy}
Moo~Jin Kim, Yihuai Gao, Tsung-Yi Lin, Yen-Chen Lin, Yunhao Ge, Grace Lam, Percy Liang, Shuran Song, Ming-Yu Liu, Chelsea Finn, et~al.
\newblock Cosmos policy: Fine-tuning video models for visuomotor control and planning.
\newblock \emph{arXiv preprint arXiv:2601.16163}, 2026{\natexlab{b}}.

\end{thebibliography}

\end{document}